\documentclass{article}

   \usepackage[final]{neurips_2021}

\usepackage[utf8]{inputenc} % allow utf-8 input
\usepackage[T1]{fontenc}    % use 8-bit T1 fonts
\usepackage{hyperref}       % hyperlinks
\usepackage{url}            % simple URL typesetting
\usepackage{booktabs}       % professional-quality tables
\usepackage{amsfonts}       % blackboard math symbols
\usepackage{nicefrac}       % compact symbols for 1/2, etc.
\usepackage{microtype}      % microtypography
\usepackage{xcolor}         % colors

\usepackage{todonotes}

\usepackage{wrapfig}

\usepackage{amsmath, amssymb, amsthm}
\usepackage{graphicx}
\usepackage{bbm}

\usepackage{pgf, tikz}
\usetikzlibrary{arrows, automata, positioning}
\tikzset{
    >=stealth',
    punkt/.style={
           rectangle,
           rounded corners,
           draw=black, very thick,
           text width=6.5em,
           minimum height=2em,
           text centered},
    pil/.style={
           ->,
           thick,
           shorten <=2pt,
           shorten >=2pt,}
}

\newcommand{\E}{\mathop{\mathbb{E}}}
\newcommand{\Prob}{\mathbb{P}}
\newcommand{\one}{\mathbbm{1}}

\usepackage{accents}

\usepackage[ruled]{algorithm2e}

\newtheorem{theorem}{Theorem}%[section]

\newtheorem{lemma}[theorem]{Lemma}

\usepackage{pifont}

\usepackage{makecell}

\usepackage{enumitem}

\usepackage{thmtools}
\usepackage{thm-restate}

\usepackage{outlines}

\usepackage[T1]{fontenc}

\usepackage{natbib}
\title{Offline RL Without Off-Policy Evaluation}
\author{%
  David Brandfonbrener \qquad William F. Whitney \qquad Rajesh Ranganath \qquad Joan Bruna\\
  Department of Computer Science, Center for Data Science\\
  New York University \\
  \texttt{david.brandfonbrener@nyu.edu} \\
}

\begin{document}

\maketitle

\begin{abstract}
    Most prior approaches to offline reinforcement learning (RL) have taken an iterative actor-critic approach involving off-policy evaluation.
    In this paper we show that simply doing one step of constrained/regularized policy improvement using an on-policy Q estimate of the behavior policy performs surprisingly well.
    This one-step algorithm beats the previously reported results of iterative algorithms on a large portion of the D4RL benchmark.
    The one-step baseline achieves this strong performance while being notably simpler and more robust to hyperparameters than previously proposed iterative algorithms.
    We argue that the relatively poor performance of iterative approaches is a result of the high variance inherent in doing off-policy evaluation and magnified by the repeated optimization of policies against those estimates. 
    In addition, we hypothesize that the strong performance of the one-step algorithm is due to a combination of favorable structure in the environment and behavior policy.
\end{abstract}

% \newpage

% \section*{Logical argument}

% \begin{outline}
% \1 One step of policy improvement performs at least as well as multiple steps
%     \2 Multiple steps of policy iteration causes Q estimation to fail
%         \3 As the policy whose value is being estimated gets further from the data, statistical issues dominate
%         \3 These issues are compounded by optimization issues that lead to Q divergence when the new policy differs too much from the behavior policy
        
%     \2 When regularized enough that Q estimation succeeds, the resulting policy is the same for one and multiple steps
%         \3 The constraint becomes binding at a smaller update than the one required to take advantage of more steps
        
%     \2 Empirically, one-step allows us to use weaker regularization without introducing statistical/optimization issues
% \end{outline}

% \newpage

\section{Introduction}

% \begin{itemize}
%     \item Offline RL/ prior work

%     \item Summary of benchmark results (guidance)

%     \item What is holding multi-step back? evaluation error caused by distribution shift and iterative error exploitation

%     \item When can multi-step be more effective? Very broad coverage to estimate bad states. When behavior is good, one-step more effective.

%     \item Summarize contributions
% \end{itemize}

An important step towards effective real-world RL is to improve sample efficiency. One avenue towards this goal is offline RL (also known as batch RL) where we attempt to learn a new policy from data collected by some other behavior policy without interacting with the environment. Recent work in offline RL is well summarized by \cite{levine2020offline}. 

In this paper, we challenge the dominant paradigm in the deep offline RL literature that primarily relies on actor-critic style algorithms that alternate between policy evaluation and policy improvement \citep{fujimoto2018off, fujimoto2019benchmarking, peng2019advantage, kumar2019stabilizing, kumar2020conservative, wang2020critic, wu2019behavior, kostrikov2021offline, jaques2019way, Siegel2020Keep, nachum2019algaedice}. All these algorithms rely heavily on off-policy evaluation to learn the critic. Instead, we find that a simple baseline which only performs one step of policy improvement using the behavior Q function often outperforms the more complicated iterative algorithms.
Explicitly, we find that our one-step algorithm beats prior results of iterative algorithms on most of the gym-mujoco \citep{brockman2016gym} and Adroit \citep{rajeswaran2017learning} tasks in the the D4RL benchmark suite \citep{fu2020d4rl}.

We then dive deeper to understand why such a simple baseline is effective. First, we examine what goes wrong for the iterative algorithms. When these algorithms struggle, it is often due to poor off-policy evaluation leading to inaccurate Q values. We attribute this to two causes: (1) distribution shift between the behavior policy and the policy to be evaluated, and (2) iterative error exploitation whereby policy optimization introduces bias and dynamic programming propagates this bias across the state space. 
We show that empirically both issues exist in the benchmark tasks and that one way to avoid these issues is to simply avoid off-policy evaluation entirely.

Finally, we recognize that while the the one-step algorithm is a strong baseline, it is not always the best choice. In the final section we provide some guidance about when iterative algorithms can perform better than the simple one-step baseline. Namely, when the dataset is large and behavior policy has good coverage of the state-action space, then off-policy evaluation can succeed and iterative algorithms can be effective. 
In contrast, if the behavior policy is already fairly good, but as a result does not have full coverage, then one-step algorithms are often preferable. 

\begin{figure}%[13]{r}{0.61\textwidth}
    %\vspace{-0.6cm}
    \centering
    \includegraphics[width=0.6\textwidth]{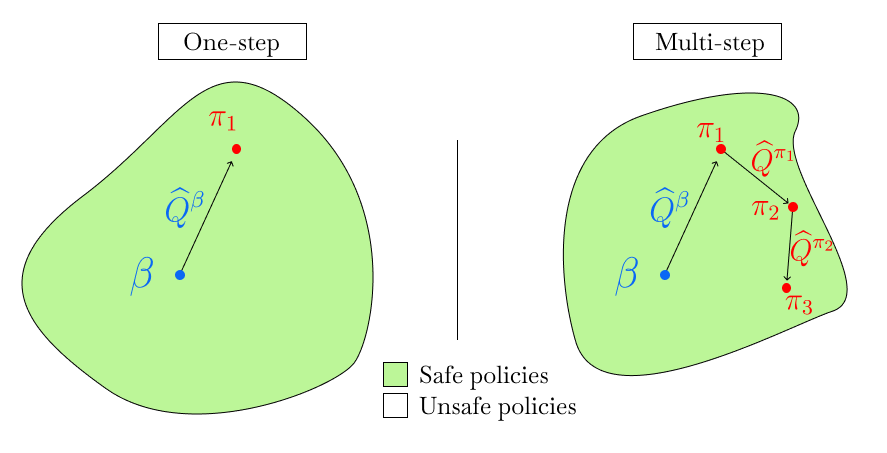}
    %\vspace{-0.2cm}
    \caption{A cartoon illustration of the difference between one-step and multi-step methods. All algorithms constrain themselves to a neighborhood of ``safe'' policies around $ \beta$. A one-step approach (left) only uses the \textcolor{blue}{on-policy} $ \widehat Q^\beta$, while a multi-step approach (right) repeatedly uses \textcolor{red}{off-policy} $\widehat Q^{\pi_i}$.
    % \joan{This is a nice illustration. I was wondering to what extent we could illustrate this neighborhood as a landscape (given by the objective function $J$) and show the two cartoon scenarios: (i) very smooth landscape, in which multi-step approaches would be worse off due to errors in estimating the gradient ($Q$), (ii) rugged landscape / valley, so that $\beta$ is in a basin that needs a non-linear path to keep descending. Also, we can illustrate what happens in off-policy updates, as the red arrows start drifting away from the true ones (not accessible)}
    }
    \label{fig:cartoon}
\end{figure}

Our main contributions are:
\begin{itemize}[leftmargin=*]
    \item A demonstration that a simple baseline of one step of policy improvement outperforms more complicated iterative algorithms on a broad set of offline RL problems.
    \item An examination of failure modes of off-policy evaluation in iterative offline RL algorithms.
    \item A description of when one-step algorithms are likely to outperform iterative approaches.
\end{itemize}

%\vspace{0.2cm}
\section{Setting and notation}

We will consider an offline RL setup as follows. Let $ \mathcal{M} = \{\mathcal{S}, \mathcal{A}, \rho, P, R, \gamma \}$ be a discounted infinite-horizon  MDP.
In this work we focus on applications in continuous control, so we will generally assume that both $ \mathcal{S}$ and $ \mathcal{A}$ are continuous and bounded. 
We consider the offline setting where rather than interacting with $\mathcal{M}$, we only have access to a dataset $ D_N$ of $ N $ tuples of $ (s_i, a_i, r_i)$ collected by some behavior policy $ \beta$ with initial state distribution $ \rho$. 
Let $ r(s,a) = \E_{r|s,a}[r]$ be the expected reward.
% Define $ d_\rho^\pi(s) := (1-\gamma) \sum_{t=0}^\infty \gamma^t \Prob_{\rho, P, \pi}(s_t = s)$ to be the discounted state visitation distribution under $ \pi$. For convenience we will often suppress $ \rho$ and just write $ d^\pi$. 
Define the state-action value function for any policy $ \pi$ by $ Q^\pi(s,a) := \E_{P, \pi | s_0 = s,\ a_0 = a}[\sum_{t=0}^\infty \gamma^t r(s_t, a_t)]$.
% We will treat samples in our dataset as iid samples from $ d^\beta_\rho $ (this assumption is potentially violated in practice, but is a common way to analyze algorithms). Throughout we will consider a statistical learning setup where we have access to some set of policies $ \Pi$ and some set of $Q$ functions $ \mathcal{Q}$. 
The objective is to maximize the expected return $ J $ of the learned policy:
\begin{align}
    J(\pi) := \E_{\rho, P, \pi}\left[ \sum_{t=0}^\infty \gamma^t r(s_t, a_t) \right] = \E_{\substack{s\sim \rho\\ a \sim \pi|s}}[Q^\pi(s,a)]. % \frac{1}{1-\gamma} \E_{s\sim d^\pi_\rho} \E_{a\sim \pi|s}[r(s,a)] 
\end{align}
Following \cite{fu2020d4rl} and others in this line of work, we allow access to the environment to tune a small (< 10) set of hyperparameters. See \cite{paine2020hyperparameter} for a discussion of the active area of research on hyperparameter tuning for offline RL. We also discuss this further in Appendix \ref{sec:app_exp_setup}.

\section{Related work}

%\subsection{The efficacy of one step of policy improvement}

%\subsection{Iterative algorithms}

\paragraph{Iterative algorithms.} Most prior work on deep offline RL consists of iterative actor-critic algorithms. The primary innovation of each paper is to propose a different mechanism to ensure that the learned policy does not stray too far from the data generated by the behavior policy. Broadly, we group these methods into three camps: policy constraints/regularization, modifications of imitation learning, and Q regularization:

\begin{enumerate}[leftmargin=*]
%\paragraph{Policy constraints/regularization.}
\item The majority of prior work acts directly on the policy. 
Some authors have proposed explicit constraints on the learned policy to only select actions where $ (s,a)$ has sufficient support under the data generating distribution \citep{fujimoto2018off, fujimoto2019benchmarking, laroche2019safe}.
%With access to $ \beta$ it is fairly trivial to implement such restriction, but can be challenging without such access. 
% These include behavior the constrained Q-learning (BCQ) algorithm and the safe policy iteration by baseline bootstrapping (SPIBB) algorithm \cite{laroche2019safe}. Both algorithms are iterative as proposed in the original work and BCQ in particular also uses a few other heuristics in its implementation like learning a policy network that modifies the proposed actions after they are sampled (and thus violating the constraints).
% These algorithms are elegant and have nice guarantees in a tabular setting. Unfortunately, they are less robust when implemented with large neural networks. This is because any errors in the Q learning or behavior estimation allow for the resulting policy to choose actions outside of the data distribution. \qfil\ avoids this problem by learning the final policy by imitation learning, thus ensuring that we imitate actions from the data distribution.
Another proposal is to regularize the learned policy towards the behavior policy \citep{wu2019behavior} usually either with a KL divergence \citep{jaques2019way} or MMD \citep{kumar2019stabilizing}. This is a very straighforward way to stay close to the behavior with a hyperparameter that determines just how close. All of these algorithms are iterative and rely on off-policy evaluation.
%These approaches can be at once too conservative by regularizing even towards the bad actions that the behavior takes and too aggressive by taking actions out of the behavior distribution if the Q function has high enough values there to outweigh the regularization. \qfil\ instead only clones the behavior actions that are high value, providing a better tradeoff between remaining within the behavior distribution while taking better actions.

%\paragraph{Variants of imitation learning.}
\item \cite{Siegel2020Keep, wang2020critic, chen2020bail} all use algorithms that filter out datapoints with low Q values and then perform imitation learning. 
%and helped to inspire \qfil. The key difference is that none of them use quantile value functions to do the filtering, so they don't have an easy to use hyperparameter to trade off bias and variance. Moreover, none of them have a safe policy improvement guarantee with function approximation.
% Perhaps most similar to our approach is the best action imitation learning (BAIL) algorithm of. 
% The algorithm attempts to learn a regularized ``upper envelope'' of the value function. The algorithm does not come with guarantees about the learned policies. The problem is that they do not do any Q-learning so are liable to eliminate data from some states entirely. Using a quantile baseline alleviates this in theory, but of course actually fitting it in practice could have the same problem if not implemented well.
\cite{wang2018exponentially, peng2019advantage} use a weighted imitation learning algorithm where the weights are determined by exponentiated Q values. 
These algorithms are iterative. 
%but since they always perform imitation learning to learn the policy, the objective always incentivizes the learned policy to stay within the support of the dataset.
%The key issue with these algorithms is that they are prone to overfitting when we have expressive models since they are not action-stable \citep{brandfonbrener2021offline}. Since the exponentiated values are always positive and every datapoint is included in the loss, the objectives proposed in these papers are perfectly optimized by choosing every action in the data with probability 1. This means that these algorithms will tend to overfit the behavior policy when given a large enough model and is closely connected to the issues of using importance weighting with neural networks raised by \cite{byrd2018effect}.

%\paragraph{Regularized Q values.}
\item Another way to prevent the learned policy from choosing unknown actions is to incorporate some form of regularization to encourage staying near the behavior and being pessimistic about unknown state, action pairs \citep{wu2019behavior, nachum2019algaedice, kumar2020conservative, kostrikov2021offline, gulcehre2021regularized}.
%The pessimism can be thought of as a softer version of a policy constraint, since higher levels of pessimism on unobserved actions make those actions less likely to be selected.
However, being able to properly quantify uncertainty about unknown states is notoriously difficult when dealing with neural network value functions \citep{buckman2020importance}. 
%\qfil\ avoids these issues by not requiring any uncertainty estimates.

\end{enumerate}

% \paragraph{Distributional RL.}
% \cite{dabney2018distributional} learns quantile Q functions for distributional RL. We are considering quantiles of a different distribution in this paper. They consider the quantiles of the distribution over future returns determined by the policy and environment along entire trajectories. We are only considering the quantiles of our estimated Q function induced by the stochasticity in the policy at only the first action.

\paragraph{One-step algorithms.} Some recent work has also noted that optimizing policies based on the behavior value function can perform surprisingly well. 
As we do, \cite{goo2020you} studies the continuous control tasks from the D4RL benchmark, but they examine a complicated algorithm involving ensembles, distributional Q functions, and a novel regularization technique. In contrast, we analyze a substantially simpler algorithm and get better performance on the D4RL tasks. We also focus more of our contribution on understanding and explaining this performance.
\cite{gulcehre2021regularized} studies the discrete action setting and finds that a one-step algorithm (which they call ``behavior value estimation'') outperforms prior work on Atari games and other discrete action tasks from the RL Unplugged benchmark \citep{gulcehre2020rl}. 
They also introduce a novel regularizer for the evaluation step. 
In contrast, we consider the continuous control setting.
This is a substantial difference in setting since continuous control requires actor-critic algorithms with parametric policies while in the discrete setting the policy improvement step can be computed exactly from the Q function. 
% The large action space means that almost all actions are not observed in the data and we must rely on generalization, potentially worsening issue of iterative error exploitation, as discussed in Section \ref{sec:why}.
Moreover, while \cite{gulcehre2021regularized} attribute the poor performance of iterative algorithms to ``overestimation'', we define and separate the issues of distribution shift and iterative error exploitation which can combine to cause overestimation. This separation helps to expose the difference between the fundamental limits of off-policy evaluation from the specific problems induced by iterative algorithms, and will hopefully be a useful distinction to inspire future work. Finally, a one-step variant is also briefly discussed in \cite{nadjahi2019safe}, but is not the focus of that work.

There are also important connections between the one-step algorithm and the literature on conservative policy improvement \citep{kakade2002approximately, schulman2015trust, achiam2017constrained}, which we discuss in more detail in Appendix \ref{sec:app_improvement}.

\section{Defining the algorithms}

In this section we provide a unified algorithmic template for model-free offline RL algorithms as offline approximate modified policy iteration. We show how this template captures our one-step algorithm as well as a multi-step policy iteration algorithm and an iterative actor-critic algorithm. 
Then any choice of policy evaluation and policy improvement operators can be used to define one-step, multi-step, and iterative algorithms. 

%Full details about our implementation can be found in Appendix \warn{write up implementation}.

\subsection{Algorithmic template}

\begin{wrapfigure}[9]{r}{0.6\textwidth}
\vspace{-0.6cm}
      \begin{algorithm}[H]
            \SetKwInOut{Input}{input}
             \Input{$ K$, dataset $ D_N$,  estimated behavior $ \hat \beta$}
             
             Set $\pi_0 = \hat \beta$. Initialize $ \widehat Q^{\pi_{-1}}$ randomly. 
             
             \For{k = 1, \dots, K}{
              Policy evaluation: $ \widehat Q^{\pi_{k-1}} = \mathcal{E}(\pi_{k-1}, D_N, \widehat Q^{\pi_{k-2}})$ 
              
              Policy improvement: $ \pi_{k} = \mathcal{I}(\widehat Q^{\pi_{k-1}}, \hat \beta, D_N, \pi_{k-1})$
             }
             \caption{OAMPI}
             \label{alg:oapi}
        \end{algorithm}
\end{wrapfigure}

We consider a generic offline approximate modified policy iteration (OAMPI) scheme, shown in Algorithm \ref{alg:oapi} (and based off of \cite{puterman1978modified, scherrer2012approximate}). Essentially the algorithm alternates between two steps. First, there is a policy evaluation step where we estimate the Q function of the current policy $ \pi_{k-1}$ by $ \widehat Q^{\pi_{k-1}}$ using only the dataset $ D_N$. Implementations also often use the prior Q estimate $ \widehat Q^{\pi_{k-2}}$ to warm-start the approximation process. Second, there is a policy improvement step. This step takes in the estimated Q function $ \widehat Q^{\pi_{k-1}}$, the estimated behavior $ \hat \beta$, and the dataset $ D_N$ and produces a new policy $ \pi_k$. Again an algorithm may use $ \pi_{k-1}$ to warm-start the optimization. Moreover, we expect this improvement step to be regularized or constrained to ensure that $ \pi_k$ remains in the support of $ \beta$ and $ D_N$. Choices for this step are discussed below. Now we discuss a few ways to instantiate the template.

% This generic template can be instantiated by several different algorithms. Our main result is that the one-step algorithm with $ K=1$ is as good or better than prior work that uses larger $ K$.

\paragraph{One-step.}
The simplest algorithm sets the number of iterations $ K = 1$. We learn $ \hat \beta$ by maximum likelihood and train the policy evaluation step to estimate $ Q^\beta$. Then we use any one of the policy improvement operators discussed below to learn $ \pi_1$. Importantly, this algorithm completely avoids off-policy evaluation.

% As the evaluation $ \mathcal{Q}$ operator we use on-policy SARSA using $ D_N$ to approximate $ Q^\beta$, trained to convergence. Then we can use any improvement operator discussed below, also trained to convergence. Explicitly the policy evaluation is done by:
% \begin{align}
%      \mathcal{Q}(\pi_0, D_N, \widehat Q^{\pi_{-1}}) = \arg\min_Q \sum_{i=1}^N (r(s_i,a_i) + \gamma Q(s_i', a_i') - Q(s_i, a_i) )^2.
% \end{align}
% In practice this is optimized by stochastic gradient descent with the use of a target network \citep{mnih2015human} to stabilize training. 

\paragraph{Multi-step.}
The multi-step algorithm now sets $ K >1$. The evaluation operator must evaluate off-policy since $ D_N$ is collected by $ \beta$, but evaluation steps for $ K \geq 2$ require evaluating policies $ \pi_{k-1}\neq \beta$. Each iteration is trained to convergence in both the estimation and improvement steps.

% There are many possible ways to do this off-policy evaluation step from simple fitted Q evaluation to various importance weighting and doubly-robust schemes. In this work, since we primarily consider continuous state and action spaces, we simple use fitted Q evaluation with no off-policy correction, following the work of \cite{fujimoto2018off, kumar2019stabilizing, Siegel2020Keep, wang2020critic} and others. Then we can again use any improvement operator discussed below and train to convergence on each iteration. Now the evaluation objective looks like 
% \begin{align}
%      \mathcal{Q}(\pi_{k-1}, D_N, \widehat Q^{\pi_{k-2}}) = \arg\min_Q \sum_{i=1}^N (r(s_i,a_i) + \gamma \E_{a'\sim \pi_{k-1}|s_i'}Q(s_i', a') - Q(s_i, a_i) )^2.
% \end{align}
% We estimate the expectation by a single sample from $ \pi_{k-1}$ and use stochastic gradient descent with a target network.

\paragraph{Iterative actor-critic.}

An actor critic approach looks somewhat like the multi-step algorithm, but does not attempt to train to convergence at each iteration and uses a much larger $ K$. Here each iteration consists of one gradient step to update the Q estimate and one gradient step to improve the policy. Since all of the evaluation and improvement operators that we consider are gradient-based, this algorithm can adapt the same evaluation and improvement operators used by the multi-step algorithm. Most algorithms from the literature fall into this category \citep{fujimoto2018off, kumar2019stabilizing, kumar2020conservative, wu2019behavior,  wang2020critic, Siegel2020Keep}.

% First, we present our simple template for one step policy improvement:

% \paragraph{Step 1: Estimate $ Q^\beta$.} We fit $ \widehat Q^\beta$ by on-policy SARSA using data from $ D_N$. 

% \paragraph{Step 2 (optional): Estimate $ \beta$.} We fit $ \hat \beta$ by maximum likelihood using data from $ D_N$. 

% \paragraph{Step 3: Policy improvement.} An algorithm then applies some policy improvement operator $ \mathcal{I}$ using the data, $ \widehat Q^\beta$, and potentially $ \hat \beta$, to produce a new policy $ \pi$.

\subsection{Policy evaluation operator}

Following prior work on continuous state and action problems, we always evaluate by simple fitted Q evaluation \citep{fujimoto2018off, kumar2019stabilizing, Siegel2020Keep, wang2020critic, paine2020hyperparameter, wang2021instabilities}.
% Explicitly the evaluation step for the one-step or multi-step algorithms looks like 
% \begin{align}
%      \mathcal{E}(\pi_{k-1}, D_N, \widehat Q^{\pi_{k-2}}) = \arg\min_Q \sum_{i=1}^N (r(s_i,a_i) + \gamma \E_{a'\sim \pi_{k-1}|s_i'}Q(s_i', a') - Q(s_i, a_i) )^2,
% \end{align}
% where the right hand side may depend on $\widehat Q^{\pi_{k-2}}$ to warm-start optimization. 
In practice this is optimized by TD-style learning with the use of a target network \citep{mnih2015human} as in DDPG \citep{lillicrap2015continuous}. 
We do not use any double Q learning or Q ensembles \citep{fujimoto2018addressing}. 
For the one-step and multi-step algorithms we train the evaluation procedure to convergence on each iteration and for the iterative algorithm each iteration takes a single stochastic gradient step.
% We estimate the expectation over next state by a single sample from $ \pi_{k-1}$ (or from the dataset in the case when $ \pi_{k-1} = \hat \beta$). 
See \cite{voloshin2019empirical, wang2021instabilities} for more comprehensive examinations of policy evaluation and some evidence that this simple fitted Q iteration approach is reasonable. It is an interesting direction for future work to consider other operators that use things like importance weighting \citep{munos2016safe} or pessimism \citep{kumar2020conservative, buckman2020importance}.

\subsection{Policy improvement operators}

To instantiate the template, we also need to choose a specific policy improvement operator $ \mathcal{I}$. We consider the following improvement operators selected from those discussed in the related work section. Each operator has a hyperparameter controlling deviation from the behavior policy.

\paragraph{Behavior cloning.} The simplest baseline worth including is to just return $ \hat \beta$ as the new policy $ \pi$. Any policy improvement operator ought to perform at least as well as this baseline.

% \paragraph{Greedy.} The naive policy improvement operator just acts greedily with respect to $ \widehat Q^\beta$ so that $\hat \pi(a|s) = \one[a = \arg\max_{a'} \widehat Q^\beta(s, a')]$.
% Calculating this arg max may not feasible in a continuous action space, so we can instead amortize the computation by learning a neural network policy to maximize the estimated Q values such that
% \begin{align}
%     \pi_k = \mathcal{I}(\widehat Q, \hat \beta, D_N, \pi_{k-1}) =  \arg\max_{\pi} \sum_i \E_{a\sim \pi|s_i} [\widehat Q (s_i, a)]
% \end{align}
% where the optimization can be performed using stochastic optimization via the reparameterization trick. Specifically, at each iteration we (1) sample a minibatch of datapoints, (2) sample one reparameterized action for each state in the minibatch, and then (3) backpropogate to get a stochastic gradient estimate.

\paragraph{Constrained policy updates.} 
Algorithms like BCQ \citep{fujimoto2018off} and SPIBB \citep{laroche2019safe} constrain the policy updates to be within the support of the data/behavior. 
% The SPIBB algorithm requires uncertainty estimates that are difficult to quantify in continuous state and action spaces, and both of these algorithms are somewhat complicated. 
In favor of simplicity, we implement a simplified version of the BCQ algorithm that removes the ``perturbation network'' which we call Easy BCQ.
We define a new policy $ \hat \pi^M_{k}$ by drawing $ M $ samples from $ \hat \beta$ and then executing the one with the highest value according to $ \widehat Q^\beta$. Explicitly:
\begin{align}
    \hat \pi^M_{k}(a|s) = \one[a = \arg\max_{a_j} \{\widehat Q^{\pi_{k-1}}(s, a_j): a_j \sim \pi_{k-1}(\cdot|s),\ 1 \leq j \leq M\}].
\end{align}

% Note, we can think about this policy in terms of pushforward distributions where we are taking an $ n$th order statistic of $ M $ samples from $ \widehat Q_\sharp^\beta (s, \hat \beta)$. If we let $ F$ be the CDF of $ \widehat Q_\sharp^\beta (s, \hat \beta)$ and $ F_M$ be the CDF of $ \widehat Q_\sharp^\beta (s, \hat \pi_M)$, it is straightforward to see that
% \begin{align}
%     F_M(v) = F(v)^M.
% \end{align}
% Thus, we know that we are moving mass in a strictly increasing way, and can come up with a policy improvement guarantee similar to the one we have for QFIL, but in terms of the difference between $ F(v) $ and $ F(v)^M$. 

% This sort of analysis is novel since prior work on BCQ \citep{fujimoto2018addressing} proposes a more complicated algorithm and only provides a tabular analysis.

\paragraph{Regularized policy updates.}
Another common idea proposed in the literature is to regularize towards the behavior policy \citep{wu2019behavior, jaques2019way, kumar2019stabilizing}. For a general divergence $ D$ we can define an algorithm that maximizes a regularized objective: 
\begin{align}
    \hat \pi^\alpha_k = \arg\max_\pi \sum_{i}  \E_{a\sim \pi|s}[\widehat Q^{\pi_{k-1}}(s_i, a)] - \alpha D(\hat \beta(\cdot|s_i), \pi(\cdot|s_i))
\end{align}

A comprehensive review of different variants of this method can be found in \cite{wu2019behavior} which does not find dramatic differences across regularization techniques. 
In practice, we will use reverse KL divergence,  i.e. $ KL( \pi(\cdot|s_i) \| \hat \beta(\cdot|s_i))$.
%For the forward KL, i.e. $ KL(\beta(\cdot|s_i) \| \pi(\cdot|s_i))$, we can simply use the sampled action $ a_i$ from the dataset to estimate the divergence. 
To compute the reverse KL, we draw samples from $ \pi(\cdot |s_i)$ and use the density estimate $ \hat \beta$ to compute the divergence. Intuitively, this regularization forces $ \pi$ to remain within the support of $ \beta$ rather than incentivizing $ \pi$ to cover $\beta$.

\paragraph{Variants of imitation learning.}
Another idea, proposed by \citep{wang2018exponentially, Siegel2020Keep, wang2020critic, chen2020bail} is to modify an imitation learning algorithm either by filtering or weighting the observed actions to incentivize policy improvement. The weighted version that we implement uses exponentiated advantage estimates to weight the observed actions:
\begin{align}
    \hat \pi^\tau_k = \arg\max_\pi \sum_{i}  \exp(\tau (\widehat Q^{\pi_{k-1}}(s_i, a_i) - \widehat V(s_i))) \log \pi(a_i|s_i).
\end{align}
With these definitions, we can now move on to testing various combinations of algorithmic template (one-step, multi-step, or iterative) and improvement operator (Easy BCQ, reverse KL regularization, or exponentially weighted imitation). 
%This can also be derived as a variational implementation of the reverse-KL regularized update, as is done in \cite{wang2018exponentially}.

% \begin{align}
%     \hat \pi_b (a|s) = \arg\max_\pi \sum_{i}  \1[\widehat Q^\beta(s_i, a_i) \geq b(s_i)] \log \pi(a_i|s_i)
% \end{align}

\section{Benchmark Results}\label{sec:bench}

Our main empirical finding is that one step of policy improvement is sufficient to beat state of the art results on much of the D4RL benchmark suite \citep{fu2020d4rl}. 
This is striking since prior work focuses on iteratively estimating the Q function of the current policy iterate, but we only use one step derived from $ \widehat Q^\beta$. Results are shown in Table \ref{tab:d4rl}. Full experimental details are in Appendix \ref{sec:app_exp_setup} and code can be found at \url{https://github.com/davidbrandfonbrener/onestep-rl}. 

\begin{table}[h]
    \centering
    \caption{Results of one-step algorithms on the D4RL benchmark. The first column gives the best results across several iterative algorithms considered in \cite{fu2020d4rl}. Each algorithm is tuned over 6 values of their respective hyperparameter. We report the mean and standard error over 10 seeds of the training process and using 100 evaluation episodes per seed. We \textbf{bold} the best result on each dataset and \textcolor{blue}{blue} any result where a one-step algorithm beat the best reported iterative result from \cite{fu2020d4rl}. We use m for medium, m-e for medium-expert, m-re for medium-replay, r for random, and c for cloned.}
    \begin{small}
    \begin{tabular}{lccccc}
        \toprule & Iterative & \multicolumn{4}{c}{One-step}\\
        \cmidrule(lr){2-2} \cmidrule(lr){3-6}
         & \cite{fu2020d4rl} & BC & Easy BCQ  & Rev. KL Reg & Exp. Weight \\
        \midrule
        halfcheetah-m & 46.3 & 
                        42.1  $\pm$  0.1  & 
                        \textcolor{blue}{52.6  $\pm$  0.1}  & 
                        \textcolor{blue}{\textbf{55.6  $\pm$  0.2 }} &  
                        \textcolor{blue}{48.6  $\pm$  0.0} \\
        walker2d-m & 81.1 & 
                        70.2  $\pm$  1.3 & 
                        \textcolor{blue}{\textbf{86.9  $\pm$  0.4 }} & 
                        \textcolor{blue}{85.6  $\pm$  0.4 } &
                        {80.3  $\pm$  1.1}  \\
        hopper-m & 58.8 & 
                        49.8  $\pm$  0.6 & 
                        \textcolor{blue}{69.7  $\pm$  2.1 }& 
                        \textcolor{blue}{\textbf{83.3  $\pm$  1.4}} & 
                        {56.7  $\pm$  0.8}\\
        \midrule               
        halfcheetah-m-e & 64.7 & 
                        60.1  $\pm$  0.8  & 
                        \textcolor{blue}{77.0  $\pm$  0.9}  &
                        \textcolor{blue}{\textbf{93.5  $\pm$  0.1}} &  
                        \textcolor{blue}{91.7  $\pm$  0.9 } \\
        walker2d-m-e & 111.0 & 
                        93.6  $\pm$  5.6 & 
                        \textcolor{blue}{111.8  $\pm$  0.2 }   &  
                        {110.9  $\pm$  0.1 }  & 
                        \textcolor{blue}{\textbf{112.9  $\pm$  0.2  }}  \\
        hopper-m-e & \textbf{111.9} & 
                        48.1  $\pm$  1.5 & 
                        81.4  $\pm$  1.9 & 
                        102.1  $\pm$  1.3 &
                        83.1  $\pm$  7.0 \\
        \midrule              
        halfcheetah-m-re & \textbf{47.7} & 
                         34.9  $\pm$  0.3  & 
                         38.4  $\pm$  0.3   &
                        42.4  $\pm$  0.1 &  
                         38.6  $\pm$  0.5 \\
        walker2d-m-re & 26.7 & 
                        23.9  $\pm$  1.6    & 
                         \textcolor{blue}{66.4  $\pm$  2.0} &  
                        \textcolor{blue}{\textbf{71.6  $\pm$  3.1}} & 
                         \textcolor{blue}{49.3  $\pm$  3.5}  \\
        hopper-m-re & 48.6 & 
                       21.2  $\pm$  1.3 & 
                        \textcolor{blue}{77.3  $\pm$  2.7 } & 
                        \textcolor{blue}{71.0  $\pm$  8.1} &
                        \textcolor{blue}{\textbf{94.1  $\pm$  2.4}}\\
        \midrule            
        halfcheetah-r & \textbf{35.4} & 
                        2.2  $\pm$  0.0 & 
                        5.4  $\pm$  0.1 & 
                        6.9  $\pm$  1.0   & 
                        3.7  $\pm$  0.2 \\
        walker2d-r & \textbf{7.3} & 
                        0.7  $\pm$  0.1 & 
                        4.2  $\pm$  0.2  & 
                        6.1  $\pm$  0.3 & 
                        5.2  $\pm$  0.2 \\
        hopper-r & \textbf{12.2} & 
                        2.6  $\pm$  0.4 & 
                        6.7  $\pm$  0.1  & 
                        7.8  $\pm$  0.3 & 
                        5.6  $\pm$  0.6   \\
        \midrule
        pen-c & 56.9 & 
                    49.3  $\pm$  2.2 & 
                    \textcolor{blue}{\textbf{67.0  $\pm$  1.1 }} &
                    {55.3  $\pm$  1.9 } &
                    {54.7  $\pm$  2.3} \\
        hammer-c & 2.1 & 
                    0.5  $\pm$  0.1  & 
                    \textcolor{blue}{\textbf{2.8  $\pm$  0.5 }} & 
                    0.2  $\pm$  0.0 & 
                    1.2  $\pm$  0.2  \\
        relocate-c & -0.1 & 
                    0.0  $\pm$  0.0 & 
                    \textcolor{blue}{\textbf{0.3  $\pm$  0.0}} & 
                    \textcolor{blue}{0.1  $\pm$  0.0} & 
                    \textcolor{blue}{0.1  $\pm$  0.0}   \\
         door-c & \textbf{0.4} & 
                    0.0  $\pm$  0.0 &
                    \textcolor{blue}{\textbf{0.4  $\pm$  0.2}} & 
                    0.0  $\pm$  0.1 & 
                    0.1  $\pm$  0.1 \\
        \bottomrule
    \end{tabular}
    \end{small}
    \label{tab:d4rl}
\end{table}

% Goo:
% m: 35.6,  68.6, 76.4
% m-e: 16.6,  92.7, 106.4,
% m-re: 34.4,  41.5, 94.8
% r: 7.2, 8.1, 9.8
% c: 61.2, 0.4, -0.2, 1.1

As we can see in the table, all of the one-step algorithms usually outperform the best iterative algorithms tested by \cite{fu2020d4rl}.
The one notable exception is the case of random data (especially on halfcheetah), where iterative algorithms have a clear advantage. We will discuss potential causes of this further in Section \ref{sec:when}.

%\begin{wraptable}[16]{r}{0.6\textwidth}
\begin{table}[h]
%\vspace{-0.8 cm}
    \centering
    \caption{Results of reverse KL regularization on the D4RL benchmark across one-step, multi-step, and iterative algorithms. Again we run 6 hyperparameters and report the mean and standard error across 10 seeds using 100 evaluation episodes.}
    \begin{small}
    \begin{tabular}{lccc}
        \toprule
         & One-step & Multi-step & Iterative \\
        \midrule
        halfcheetah-m & 
                        \textbf{55.6  $\pm$  0.2}  &  
                        40.8  $\pm$  8.6 &
                        47.4  $\pm$  3.5\\
        walker2d-m & 
                        \textbf{85.6  $\pm$  0.4} &
                        75.9  $\pm$  0.5    & 
                        75.4  $\pm$  0.8  \\
        hopper-m & 
                        \textbf{83.3  $\pm$  1.4} & 
                        53.0  $\pm$  1.0  & 
                        54.2  $\pm$  0.6 \\
        \midrule
        halfcheetah-m-e &
                        \textbf{93.5  $\pm$  0.1}  &  
                        \textbf{93.6  $\pm$  0.3}  & 
                        \textbf{93.6  $\pm$  0.2}  \\
        walker2d-m-e & 
                        \textbf{110.9  $\pm$  0.1} & 
                        76.3  $\pm$  15.9  & 
                        108.2  $\pm$  0.3  \\
        hopper-m-e & 
                        \textbf{102.1  $\pm$  1.3 } &
                        \textbf{101.3  $\pm$  3.9} & 
                        82.7  $\pm$  7.4 \\
        \midrule          
        halfcheetah-r &  
                        6.9  $\pm$  1.0  & 
                        13.7  $\pm$  1.7 & 
                        \textbf{16.3  $\pm$  1.6}\\
        walker2d-r & 
                        \textbf{6.1  $\pm$  0.3} & 
                        5.0  $\pm$  0.3  &
                        5.1  $\pm$  0.3 \\
        hopper-r & 
                        7.8  $\pm$  0.3 & 
                        \textbf{15.4  $\pm$  2.9} & 
                        9.7  $\pm$  0.1\\
        \bottomrule
    \end{tabular}
    \end{small}
    \label{tab:multi}
\end{table}

To give a more direct comparison that controls for any potential implementation details, we use our implementation of reverse KL regularization to create multi-step and iterative algorithms. We are not using algorithmic modifications like Q ensembles, regularized Q values, or early stopping that have been used in prior work. But, our iterative algorithm recovers similar performance to prior regularized actor-critic approaches. These results are shown in Table \ref{tab:multi}.

Put together, these results immediately suggest some guidance to the practitioner: it is worthwhile to run the one-step algorithm as a baseline before trying something more elaborate. The one-step algorithm is substantially simpler than prior work, but frequently achieves better performance.

% \subsection{Benchmarking QFIL}

% \begin{figure}[h]
%     \centering
%     \includegraphics[width=0.3\textwidth]{figures/benchmark/hc-m.png}\includegraphics[width=0.3\textwidth]{figures/benchmark/hc-me.png}\includegraphics[width=0.3\textwidth]{figures/benchmark/hc-r.png}\\
%     \includegraphics[width=0.3\textwidth]{figures/benchmark/w-m.png}\includegraphics[width=0.3\textwidth]{figures/benchmark/w-me.png}\includegraphics[width=0.3\textwidth]{figures/benchmark/w-r.png}\\
%     \includegraphics[width=0.3\textwidth]{figures/benchmark/h-m.png}\includegraphics[width=0.3\textwidth]{figures/benchmark/h-me.png}\includegraphics[width=0.3\textwidth]{figures/benchmark/h-r.png}
%     \caption{Results on the D4RL benchmark, with state of the art (SOTA) taken from \cite{fu2020d4rl}. \qfil\ beats the state of the art on 6 out of 9 datasets shown in the figure and always makes a substantial improvement over behavior cloning (BC).}
%     \label{fig:benchmark}
% \end{figure}

\section{What goes wrong for iterative algorithms?}\label{sec:why}

% \subsection{Policy improvement perspective}

% We can modify the performance difference lemma to measure the difference between two policies $ \pi_1, \pi_2$ using the state distribution derived from the behavior $ \beta$. 

% \begin{lemma}[Approximate three-way performance difference]
% For any three policies $ \pi_2, \pi_1$ and $ \beta$,
% \begin{align}
%     (1-\gamma) (J(\pi_2) &- J(\pi_1)) = \underbrace{\E_{s\sim d_\beta}[\widehat Q^{\pi_1}(s,\pi_2) - \widehat Q^{\pi_1}(s,{\pi_1})]}_{Estimated\ improvement} \\ &- \underbrace{\E_{s\sim d_\beta}\left[(\widehat Q^{\pi_1}(s,\pi_2) - Q^{\pi_1}(s, \pi_2)) + (Q^{\pi_1}(s,{\pi_1}) - \widehat Q^{\pi_1}(s, {\pi_1})) \right]}_{Evaluation\ error}\\
%     &- \underbrace{\E_{\substack{s\sim d_\beta \\ s' \sim d_{\pi_2}}}\left[(Q^{\pi_1}(s, \pi_2) - Q^{\pi_1}(s', \pi_2) ) + (Q^{\pi_1}(s', \pi_1) - Q^{\pi_1}(s, \pi_1)) \right]}_{State\ distribution\ mismatch}
% \end{align}
% \end{lemma}

The benchmark experiments show that one step of policy improvement often beats iterative and multi-step algorithms. In this section we dive deeper to understand why this happens. First, by examining the learning curves of each of the algorithms we note that iterative algorithms require stronger regularization to avoid instability. Then we identify two causes of this instability: \emph{distribution shift} and \emph{iterative error exploitation}. 

Distribution shift causes evaluation error by reducing the effective sample size in the fixed dataset for evaluating the current policy and has been extensively considered in prior work as discussed below. Iterative error exploitation occurs when we repeatedly optimize policies against our Q estimates and exploit their errors. This introduces a bias towards overestimation at each step (much like the training error in supervised learning is biased to be lower than the test error). Moreover, by iteratively re-using the data and using prior Q estimates to warmstart training at each step, the errors from one step are amplified at the next. This type of error is particular to multi-step and iterative algorithms.

\subsection{Learning curves and hyperparameter sensitivity}

\begin{figure}[h]
    \centering
    \includegraphics[width=\textwidth]{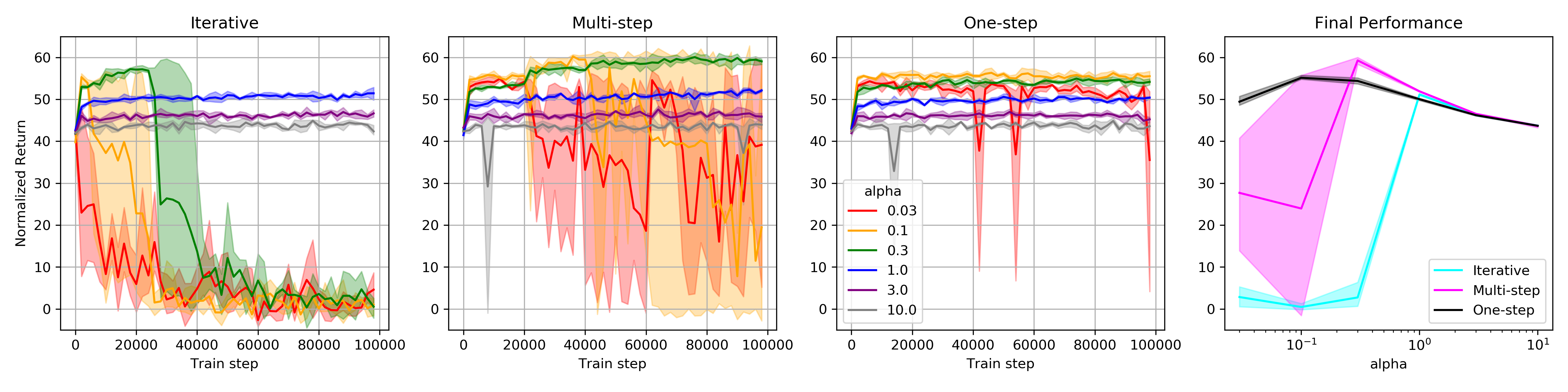}
    \caption{Learning curves and final performance on halfcheetah-medium across different algorithms and regularization hyperparameters (all using the reverse KL regularized improvement operator). Error bars show min and max over 3 seeds. Similar figures for other datasets from D4RL can be found in Appendix \ref{sec:app_extra_exp}.}
    \label{fig:learning_curves}
\end{figure}

To begin to understand why iterative and multi-step algorithms can fail it is instructive to look at the learning curves. As shown in Figure \ref{fig:learning_curves}, we often observe that the iterative algorithm will begin to learn and then crash. Regularization can help to prevent this crash since strong enough regularization towards the behavior policy ensures that the evaluation is nearly on-policy. 

In contrast, the one-step algorithm is more robust to the regularization hyperparameter. The rightmost panel of the figure shows this clearly. While iterative and multi-step algorithms can have their performance degrade very rapidly with the wrong setting of the hyperparameter, the one-step approach is more stable.
Moreover, we usually find that the optimal setting of the regularization hyperparameter is lower for the one-step algorithm than the iterative or multi-step approaches.

\subsection{Distribution shift}

Any algorithm that relies on off-policy evaluation will struggle with distribution shift in the evaluation step. Trying to evaluate a policy that is substantially different from the behavior reduces the effective sample size and increases the variance of the estimates. Explicitly, by distribution shift we mean the shift between the behavior distribution (the distribution over state-action pairs in the dataset) and the evaluation distribution (the distribution that would be induced by the policy $ \pi$ we want to evaluate).

\paragraph{Prior work.} There is a substantial body of prior theoretical work that suggests that off-policy evaluation can be difficult and this difficulty scales with some measure of distribution shift. 
%\cite{jiang2016doubly} give an exponential (in horizon) lower bound on sample complexity in the tabular setting for any unbiased estimator from the Cramer-Rao lower bound. 
\cite{wang2020statistical, Amortila2020AVO, zanette2021exponential} give exponential (in horizon) lower bounds on sample complexity in the linear setting even with good feature representations that can represent the desired Q function and assuming good data coverage. 
Upper bounds generally require very strong assumptions on both the representation and limits on the distribution shift \citep{wang2021instabilities, duan2020minimax, chen2019information}. Moreover, the assumed bounds on distribution shift can be exponential in horizon in the worst case.
% The literature on policy improvement also suggests limiting distribution shift \citep{kakade2002approximately, schulman2015trust, achiam2017constrained}. We discuss how the guarantees from this literature apply directly to the one-step but not multi-step algorithms in more detail in Appendix \ref{sec:app_improvement}. \warn{TODO: link this in better}
On the empirical side, \cite{wang2021instabilities} demonstrates issues with distribution shift when learning from pre-trained features and provides a nice discussion of why distribution shift causes error amplification. \cite{fujimoto2018off} raises a similar issue under the name ``extrapolation error''. Regularization and constraints are meant to reduce issues stemming from distribution shift, but also reduce the potential for improvement over the behavior.

\begin{figure}[h]
    \centering
    \includegraphics[width=\textwidth]{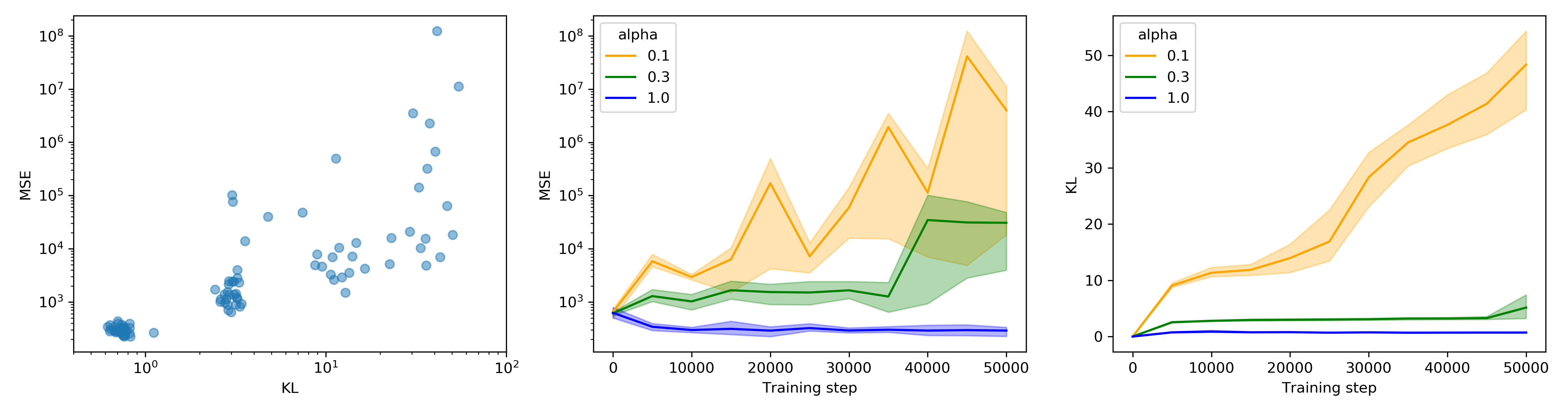}
    \caption{Results of running the iterative algorithm on halfcheetah-medium. Each checkpointed policy is evaluated by a Q function trained from scratch on heldout data. MSE refers to $\E_{s,a\sim \beta}[(\hat Q^{\pi_i}(s,a) - Q^{\pi_i}(s,a))^2]$ and KL refers to $ \E_{s\sim \beta}[KL(\pi(\cdot|s)\| \beta(\cdot|s)]$. Left: 90 policies taken from various points in training with various hyperaparmeters and random seeds. Center: MSE learning curves. Right: KL learning curves. Error bars show min and max over 3 random seeds.}
    \label{fig:mse}
\end{figure}

\paragraph{Empirical evidence.} Both the multi-step and iterative algorithms in our experiments rely on off-policy evaluation as a key subroutine. We examine how easy it is to evaluate the policies encountered along the learning trajectory. To control for issues of iterative error exploitation (discussed in the next subsection), we train Q estimators from scratch on a heldout evaluation dataset sampled from the behavior policy. We then evaluate these trained Q function on rollouts from 1000 datapoints sampled from the replay buffer. Results are shown in Figure \ref{fig:mse}.

The results show a correlation betweed KL and MSE. Moreover, we see that the MSE generally increases over training. One way to mitigate this, as seen in the figure, is to use a large value of $ \alpha$. We just cannot take a very large step before running into problems with distribution shift. But, when we take such a small step, the information from the on-policy $ \widehat Q^\beta$ is about as useful as the newly estimated $ \widehat Q^\pi$. This is seen, for example, in Figure \ref{fig:learning_curves} where we get very similar performance across algorithms at high levels of regularization.

\subsection{Iterative error exploitation}

The previous subsection identifies how any algorithm that uses off-policy evaluation is fundamentally limited by distribution shift, even if we were given fresh data and trained Q functions from scratch at every iteration. But, in practice, iterative algorithms repeatedly iterate between optimizing policies against estimated Q functions and re-estimating the Q functions using the \emph{same data} and using the Q function from the previous step to warm-start the re-estimation. This induces dependence between steps that causes a problem that we call iterative error exploitation.

\paragraph{Intuition about the problem.} In short, iterative error exploitation happens because $ \pi_i$ tends to choose overestimated actions in the policy improvement step, and then this overestimation propagates via dynamic programming in the policy evaluation step. To illustrate this issue more formally, consider the following: at each $ s,a$ we suffer some Bellman error $ \varepsilon_\beta^\pi(s,a)$ based on our fixed dataset collected by $ \beta$. Formally,
\begin{align}
    \widehat Q^\pi(s,a) = r(s,a) + \gamma \E_{\substack{s'|s,a \\ a'\sim \pi|s'}}[\widehat Q^\pi(s',a')] + \varepsilon_\beta^\pi(s,a).
\end{align}
Intuitively, $ \varepsilon_\beta^\pi$ will be larger at state-actions with less coverage in the dataset collected by $ \beta$. Note that $ \varepsilon_\beta^\pi $ can absorb all error whether it is caused by the finite sample size or function approximation error.

%So far this is all fully general, such an $ \varepsilon_\beta^\pi$ can always be defined for any estimated Q function. Now we will make the simplifying assumption. 
All that is needed to cause iterative error exploitation is that the $ \epsilon_\beta^\pi$ are highly correlated across different $ \pi$, but for simplicity, we will assume that $ \varepsilon_\beta^\pi$ is \emph{the same} for all policies $ \pi$ estimated from our fixed offline dataset and instead write $ \varepsilon_\beta$. Now that the errors do not depend on the policy we can treat the errors as auxiliary rewards that obscure the true rewards and see that 
\begin{align}
    \widehat Q^\pi(s,a) = Q^\pi(s,a) + \widetilde Q^\pi_\beta(s,a), \qquad \widetilde Q^\pi_\beta(s,a) := \E_{\pi| s_0, a_0 = s,a }\left[\sum_{t=0}^\infty \gamma^t \varepsilon_\beta(s_t,a_t) \right].
\end{align}
This assumption is somewhat reasonable since we expect the error to primarily depend on the data. And, when the prior Q function is used to warm-start the current one (as is generally the case in practice), the approximation errors are automatically passed between steps.

Now we can explain the problem. Recall that under our assumption the $ \varepsilon_\beta $ are fixed once we have a dataset and likely to have larger magnitude the further we go from the support of the dataset. So, with each step $ \pi_i$ is able to better maximize $ \varepsilon_\beta$, thus moving further from $ \beta$ and increasing the magnitude of $ \widetilde Q^{\pi_i}_\beta$ relative to $ Q^{\pi_i}$. Even though $ Q^{\pi_i}$ may provide better signal than $ Q^\beta$, it can easily be drowned out by $ \widetilde Q^{\pi_i}_\beta$. In contrast, $ \widetilde Q_\beta^\beta$ has small magnitude, so the one-step algorithm is robust to errors\footnote{We should note that iterative error exploitation is similar to the overestimation addressed by double Q learning \citep{van2016deep, fujimoto2018addressing}, but distinct. Since we are in the offline setting, the errors due to our finite dataset can be iteratively exploited more and more, while in the online setting considered by double Q learning, fresh data prevents this issue. We are also considering an algorithm based on policy iteration rather than value iteration.}.

\begin{figure}[h]
    \centering
    \includegraphics[width=\textwidth]{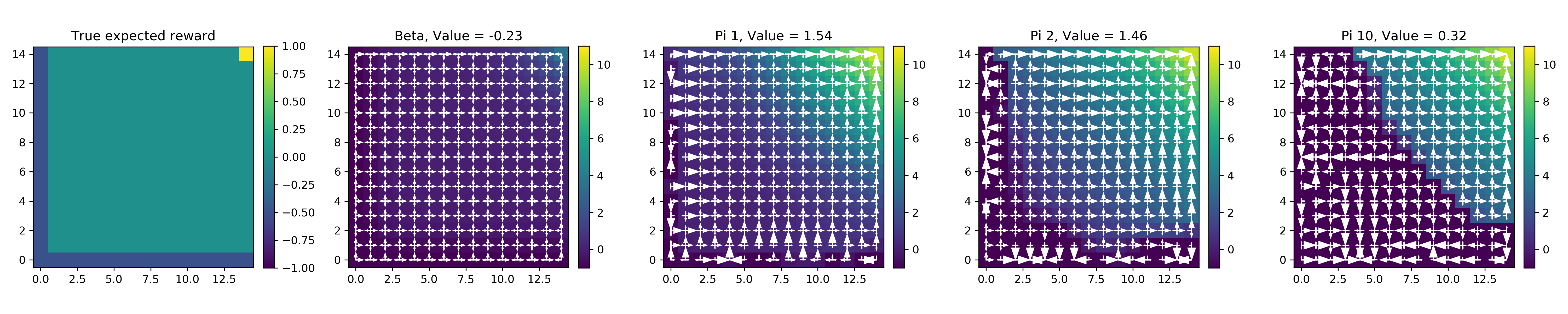}
    \includegraphics[width=\textwidth]{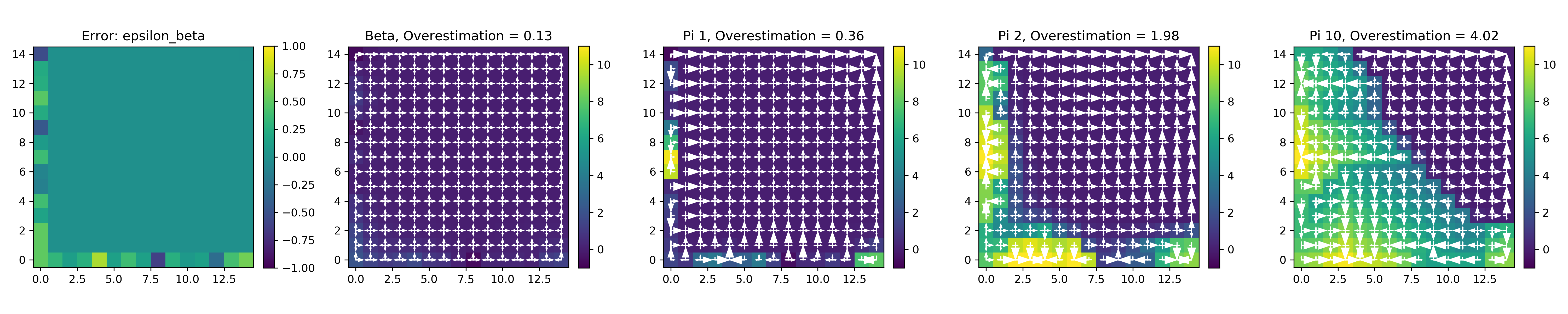}
    \caption{An illustration of multi-step offline regularized policy iteration. The leftmost panel in each row shows the true reward (top) or error $ \varepsilon_\beta$ (bottom). Then each subsequent panel plots $ \pi_i$ (with arrow size proportional to $ \pi_i(a|s)$) over either $ Q^{\pi_i}$ (top) or $ \widetilde Q^{\pi}_\beta $ (bottom), averaged over actions at each state. The one-step policy ($ \pi_1$) has the highest value. The behavior policy here is a mixture of optimal $ \pi^*$ and uniform $ u $ with coefficient 0.2 so that $\beta = 0.2 \cdot \pi^* + 0.8 \cdot u$. We set $ \alpha = 0.1$ as the regularization parameter for reverse KL regularization.}
    \label{fig:gridworld}
\end{figure}

\paragraph{An example.} Now we consider a simple gridworld example to illustrate iterative error exploitation. This example fits exactly into the setup outlined above since all errors are due to reward estimation so the $ \varepsilon_\beta$ is indeed constant over all $ \pi$. The gridworld we consider has one deterministic good state with reward 1 and many stochastic bad states that have rewards distributed as $ \mathcal{N}(-0.5, 1)$. We collect a dataset of 100 trajectories, each of length 100. One run of the multi-step offline regularized policy iteration algorithm is illustrated in Figure \ref{fig:gridworld}.

In the example we see that one step often outperforms multiple steps of improvement. Intuitively, when there are so many noisy states, it is likely that a few of them will be overestimated. Since the data is re-used for each step, these overestimations persist and propagate across the state space due to iterative error exploitation. This property of having many bad, but poorly estimated states likely also exists in the high-dimensional control problems encountered in the benchmark where there are many ways for the robots to fall down that are not observed in the data for non-random behavior. 
%\begin{wrapfigure}[12]{r}{0.61\textwidth}
%\vspace{-0.5cm}
Moreover, both settings have larger errors in areas where we have less data.
So even though the errors in the gridworld are caused by noise in the rewards, while errors in D4RL are caused by function approximation, we think this is a useful mental model of the problem.

\begin{figure}[h]
    \centering
    \includegraphics[width=0.6\textwidth]{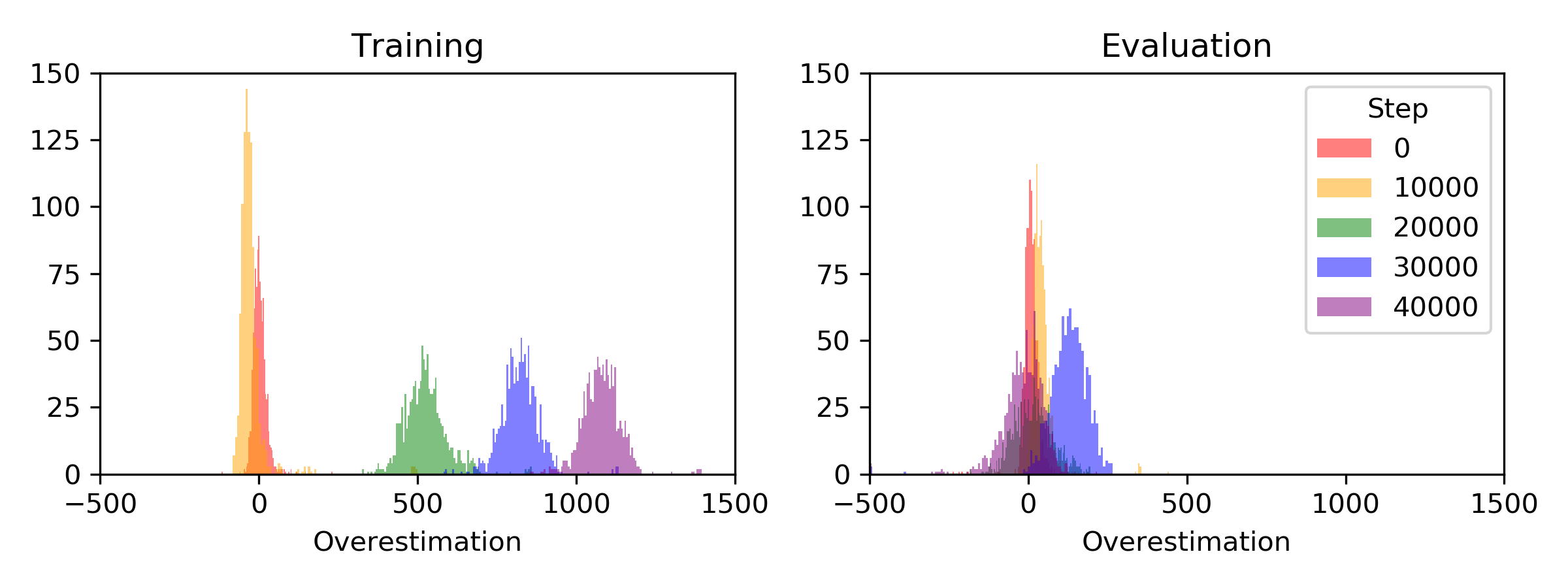}
    \caption{Histograms of overestimation error ($\widehat Q^{\pi_i}(s,a) - Q^{\pi_i}(s,a)$) on halfcheetah-medium with the iterative algorithm. Left: errors from the training Q function. Right: errors from an independently trained Q function.}
    \label{fig:over}
\end{figure}

\paragraph{Empirical evidence.}
In practice we cannot easily visualize the progression of errors. However, the dependence between steps still arises as overestimation of the Q values. We can track the overestimation of the Q values over training as a way to measure how much bias is being induced by optimizing against our dependent Q estimators. As a control we can also train Q estimators from scratch on independently sampled evaluation data. These independently trained Q functions do not have the same overestimation bias even though the squared error does tend to increase as the policy moves further from the behavior (as seen in Figure \ref{fig:mse}).
Explicitly, we track 1000 state, action pairs from the replay buffer over training. For each checkpointed policy we perform 3 rollouts at each state to get an estimate of the true Q value and compare this to the estimated Q value. Results are shown in Figure \ref{fig:over}.

\section{When are multiple steps useful?}\label{sec:when}

So far we have focused on why the one-step algorithm often works better than the multi-step and iterative algorithms. However, we do not want to give the impression that one-step is always better. Indeed, our own experiments in Section \ref{sec:bench} show a clear advantage for the multi-step and iterative approaches when we have randomly collected data. While we cannot offer a precise delineation of when one-step will outperform multi-step, in this section we offer some intuition as to when we can expect to see benefits from multiple steps of policy improvement. 

As seen in Section \ref{sec:why}, multi-step and iterative algorithms have problems when they propagate estimation errors. This is especially problematic in noisy and/or high dimensional environments. While the multi-step algorithms propagate this noise more widely than the one-step algorithm, they also propagate the signal. So, when we have sufficient coverage to reduce the magnitude of the noise, this increased propagation of signal can be beneficial. The D4RL experiments suggest that we are usually on the side of the tradeoff where the errors are large enough to make one-step preferable.

% Formally, we can think about propagation as follows. Define $ d_\pi(s', a'|s,a) = (1-\gamma) \sum_{t=0}^\infty \gamma^t P^\pi(s_t, a_t = s',a'| s_0, a_0 = s,a)$ to be the discounted future probability of visiting $ s', a'$ when executing $ \pi$ starting from $ s,a$. Then 
% \begin{align}
%     Q^{\pi_i}(s,a) = \sum_{s',a'} d_{\pi_i}(s', a'|s,a) \cdot r(s', a')
% \end{align}

%\begin{wrapfigure}[16]{r}{0.41\textwidth}
%\vspace{-0.3cm}
\begin{figure}
    \centering
    \includegraphics[width=0.4\textwidth]{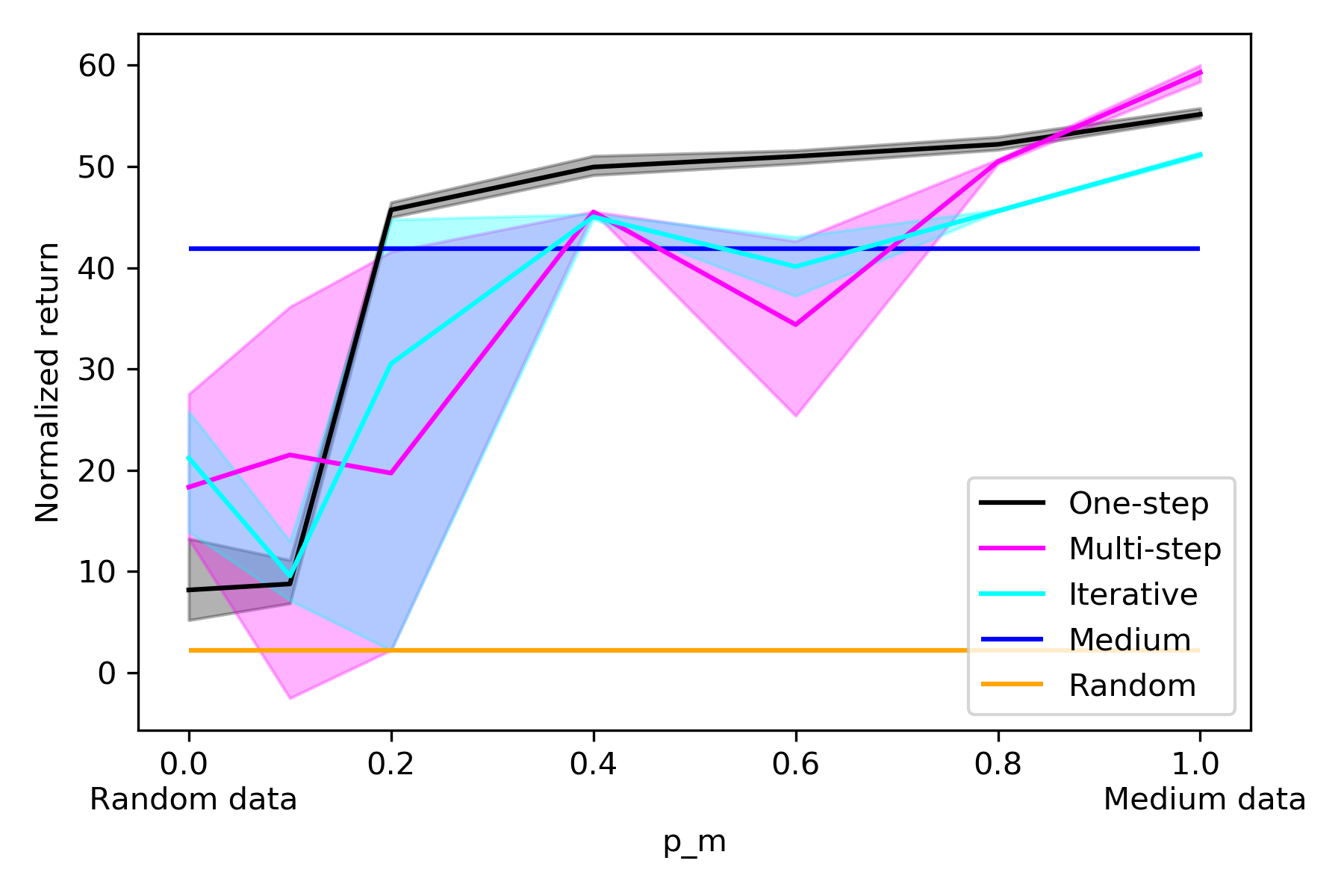}
    \caption{Performance of all three algorithms with reverse KL regularization across mixtures between halfcheetah-random and halfcheetah-medium. Error bars indicate min and max over 3 seeds.}
    \label{fig:interp}
\end{figure}

In Appendix \ref{sec:app_grid} we illustrate a simple gridworld example where a slight modification of the behavior policy from Figure \ref{fig:gridworld} makes multi-step dramatically outperform one-step. This modified behavior policy (1) has better coverage of the noisy states (which reduces error, helping multi-step), and (2) does a worse job propagating the reward from the good state (hurting one-step).

We can also test empirically how the behavior policy effects the tradeoff between error and signal propagation. To do this we construct a simple experiment where we mix data from the random behavior policy with data from the medium behavior policy. Explicitly we construct a dataset $ D $ out of the datasets $ D_r$ for random and $ D_m$ for medium such that each trajectory in $ D $ comes from the medium dataset with probability $ p_m$. So for $ p_m = 0$ we have the random dataset and $ p_m = 1 $ we have the medium dataset, and in between we have various mixtures. Results are shown in Figure \ref{fig:interp}. It takes surprisingly little data from the medium policy for one-step to outperform the iterative algorithm.

\section{Discussion, limitations, and future work}

This paper presents the surprising effectiveness of a simple one-step baseline for offline RL. We examine the failure modes of iterative algorithms and the conditions where we might expect them to outperform the simple one-step baseline. This provides guidance to a practitioner that the simple one-step baseline is a good place to start when approaching an offline RL problem.

But, we leave many questions unanswered. 
One main limitation is that we lack a clear theoretical characterization of which environments and behaviors can guarantee that one-step outperforms multi-step or visa versa. Such results will likely require strong assumptions, but could provide useful insight. We don't expect this to be easy as it requires understanding policy iteration which has been notoriously difficult to analyze, often converging much faster than the theory would suggest \citep{sutton2018reinforcement, Agarwal2019ReinforcementLT}.
Another limitation is that while only using one step is perhaps the simplest way to avoid the problems of off-policy evaluation, there are possibly other more elaborate algorithmic solutions that we did not consider here. However, our strong empirical results suggest that the one-step algorithm is at least a strong baseline.
% Some final questions we would like to leave for the reader are: how much improvement over the behavior can we expect from offline RL? Is there a way to characterize this a priori in simplified settings? Or an efficient way to test how much to trust a learned policy without interacting with the environment?

\paragraph{Broader impact.} Our paper studies a simple and effective baseline approach to the offline RL problem. The effectiveness of this baseline raises some serious questions about the utility of prior work proposing substantially more complicated methods. By making this observation of prior shortcomings, our paper has the potential to encourage researchers to derive new and better methods for offline RL. This has many potential impacts on fields as diverse as robotics and healthcare where better offline decision making can lead to better real-world performance. As always, we note that machine learning improvements come in the form of “building machines to do X better”. For a sufficiently malicious or ill-informed choice of X, almost any progress in machine learning might indirectly lead to a negative outcome, and our work is not excluded from that.

\subsection*{Acknowledgements}
This work is partially supported by the Alfred P. Sloan Foundation, NSF RI-1816753, NSF CAREER CIF 1845360, NSF CHS-1901091, Samsung Electronics, and the Institute for Advanced Study.
DB is supported by the Department of Defense (DoD) through the National Defense Science \& Engineering Graduate Fellowship (NDSEG) Program.

\bibliography{rl.bib}

\begin{thebibliography}{45}
\providecommand{\natexlab}[1]{#1}
\providecommand{\url}[1]{\texttt{#1}}
\expandafter\ifx\csname urlstyle\endcsname\relax
  \providecommand{\doi}[1]{doi: #1}\else
  \providecommand{\doi}{doi: \begingroup \urlstyle{rm}\Url}\fi

\bibitem[Achiam et~al.(2017)Achiam, Held, Tamar, and
  Abbeel]{achiam2017constrained}
Joshua Achiam, David Held, Aviv Tamar, and Pieter Abbeel.
\newblock Constrained policy optimization.
\newblock In \emph{International Conference on Machine Learning}, pages 22--31.
  PMLR, 2017.

\bibitem[Agarwal et~al.(2019)Agarwal, Jiang, and
  Kakade]{Agarwal2019ReinforcementLT}
Alekh Agarwal, Nan Jiang, and S.~Kakade.
\newblock Reinforcement learning: Theory and algorithms.
\newblock 2019.

\bibitem[Amortila et~al.(2020)Amortila, Jiang, and Xie]{Amortila2020AVO}
P.~Amortila, Nan Jiang, and Tengyang Xie.
\newblock A variant of the wang-foster-kakade lower bound for the discounted
  setting.
\newblock \emph{ArXiv}, abs/2011.01075, 2020.

\bibitem[Brockman et~al.(2016)Brockman, Cheung, Pettersson, Schneider,
  Schulman, Tang, and Zaremba]{brockman2016gym}
Greg Brockman, Vicki Cheung, Ludwig Pettersson, Jonas Schneider, John Schulman,
  Jie Tang, and Wojciech Zaremba.
\newblock Openai gym.
\newblock \emph{CoRR}, abs/1606.01540, 2016.
\newblock URL \url{http://arxiv.org/abs/1606.01540}.

\bibitem[Buckman et~al.(2020)Buckman, Gelada, and
  Bellemare]{buckman2020importance}
Jacob Buckman, Carles Gelada, and Marc~G. Bellemare.
\newblock The importance of pessimism in fixed-dataset policy optimization,
  2020.

\bibitem[Burkhardt(2014)]{Burkhardt2014truncated}
John Burkhardt.
\newblock The truncated normal distribution, 2014.

\bibitem[Chen and Jiang(2019)]{chen2019information}
Jinglin Chen and Nan Jiang.
\newblock Information-theoretic considerations in batch reinforcement learning.
\newblock In \emph{Proceedings of the 36th International Conference on Machine
  Learning}. PMLR, 2019.

\bibitem[Chen et~al.(2020)Chen, Zhou, Wang, Wang, Wu, and Ross]{chen2020bail}
Xinyue Chen, Zijian Zhou, Zheng Wang, Che Wang, Yanqiu Wu, and Keith Ross.
\newblock Bail: Best-action imitation learning for batch deep reinforcement
  learning.
\newblock \emph{Advances in Neural Information Processing Systems}, 33, 2020.

\bibitem[Duan et~al.(2020)Duan, Jia, and Wang]{duan2020minimax}
Yaqi Duan, Zeyu Jia, and Mengdi Wang.
\newblock Minimax-optimal off-policy evaluation with linear function
  approximation.
\newblock In \emph{International Conference on Machine Learning}, pages
  2701--2709. PMLR, 2020.

\bibitem[Fu et~al.(2020)Fu, Kumar, Nachum, Tucker, and Levine]{fu2020d4rl}
Justin Fu, Aviral Kumar, Ofir Nachum, George Tucker, and Sergey Levine.
\newblock D4rl: Datasets for deep data-driven reinforcement learning.
\newblock \emph{arXiv preprint arXiv:2004.07219}, 2020.

\bibitem[Fujimoto et~al.(2018{\natexlab{a}})Fujimoto, Meger, and
  Precup]{fujimoto2018off}
Scott Fujimoto, David Meger, and Doina Precup.
\newblock Off-policy deep reinforcement learning without exploration.
\newblock \emph{arXiv preprint arXiv:1812.02900}, 2018{\natexlab{a}}.

\bibitem[Fujimoto et~al.(2018{\natexlab{b}})Fujimoto, van Hoof, and
  Meger]{fujimoto2018addressing}
Scott Fujimoto, Herke van Hoof, and David Meger.
\newblock Addressing function approximation error in actor-critic methods.
\newblock \emph{arXiv preprint arXiv:1802.09477}, 2018{\natexlab{b}}.

\bibitem[Fujimoto et~al.(2019)Fujimoto, Conti, Ghavamzadeh, and
  Pineau]{fujimoto2019benchmarking}
Scott Fujimoto, Edoardo Conti, Mohammad Ghavamzadeh, and Joelle Pineau.
\newblock Benchmarking batch deep reinforcement learning algorithms.
\newblock \emph{arXiv preprint arXiv:1910.01708}, 2019.

\bibitem[Goo and Niekum(2020)]{goo2020you}
Wonjoon Goo and Scott Niekum.
\newblock You only evaluate once – a simple baseline algorithm for offline
  rl.
\newblock In \emph{Offline Reinforcement Learning Workshop at Neural
  Information Processing Systems}, 2020.

\bibitem[Gulcehre et~al.(2020)Gulcehre, Wang, Novikov, Paine, Colmenarejo,
  Zolna, Agarwal, Merel, Mankowitz, Paduraru, et~al.]{gulcehre2020rl}
Caglar Gulcehre, Ziyu Wang, Alexander Novikov, Tom~Le Paine, Sergio~G{\'o}mez
  Colmenarejo, Konrad Zolna, Rishabh Agarwal, Josh Merel, Daniel Mankowitz,
  Cosmin Paduraru, et~al.
\newblock Rl unplugged: Benchmarks for offline reinforcement learning.
\newblock \emph{arXiv preprint arXiv:2006.13888}, 2020.

\bibitem[Gulcehre et~al.(2021)Gulcehre, Colmenarejo, Wang, Sygnowski, Paine,
  Zolna, Chen, Hoffman, Pascanu, and de~Freitas]{gulcehre2021regularized}
Caglar Gulcehre, Sergio~G{\'o}mez Colmenarejo, Ziyu Wang, Jakub Sygnowski,
  Thomas Paine, Konrad Zolna, Yutian Chen, Matthew Hoffman, Razvan Pascanu, and
  Nando de~Freitas.
\newblock Regularized behavior value estimation.
\newblock \emph{arXiv preprint arXiv:2103.09575}, 2021.

\bibitem[Jaques et~al.(2019)Jaques, Ghandeharioun, Shen, Ferguson, Lapedriza,
  Jones, Gu, and Picard]{jaques2019way}
Natasha Jaques, Asma Ghandeharioun, Judy~Hanwen Shen, Craig Ferguson, Agata
  Lapedriza, Noah Jones, Shixiang Gu, and Rosalind Picard.
\newblock Way off-policy batch deep reinforcement learning of implicit human
  preferences in dialog, 2019.

\bibitem[Kakade and Langford(2002)]{kakade2002approximately}
Sham Kakade and John Langford.
\newblock Approximately optimal approximate reinforcement learning.
\newblock In \emph{ICML}, volume~2, pages 267--274, 2002.

\bibitem[Kingma and Ba(2014)]{kingma2014adam}
Diederik~P Kingma and Jimmy Ba.
\newblock Adam: A method for stochastic optimization.
\newblock \emph{arXiv preprint arXiv:1412.6980}, 2014.

\bibitem[Kostrikov et~al.(2021)Kostrikov, Tompson, Fergus, and
  Nachum]{kostrikov2021offline}
Ilya Kostrikov, Jonathan Tompson, Rob Fergus, and Ofir Nachum.
\newblock Offline reinforcement learning with fisher divergence critic
  regularization.
\newblock \emph{arXiv preprint arXiv:2103.08050}, 2021.

\bibitem[Kumar et~al.(2019)Kumar, Fu, Soh, Tucker, and
  Levine]{kumar2019stabilizing}
Aviral Kumar, Justin Fu, Matthew Soh, George Tucker, and Sergey Levine.
\newblock Stabilizing off-policy q-learning via bootstrapping error reduction.
\newblock In \emph{Advances in Neural Information Processing Systems}, pages
  11761--11771, 2019.

\bibitem[Kumar et~al.(2020)Kumar, Zhou, Tucker, and
  Levine]{kumar2020conservative}
Aviral Kumar, Aurick Zhou, George Tucker, and Sergey Levine.
\newblock Conservative q-learning for offline reinforcement learning.
\newblock \emph{arXiv preprint arXiv:2006.04779}, 2020.

\bibitem[Laroche et~al.(2019)Laroche, Trichelair, and
  Des~Combes]{laroche2019safe}
Romain Laroche, Paul Trichelair, and Remi~Tachet Des~Combes.
\newblock Safe policy improvement with baseline bootstrapping.
\newblock In \emph{International Conference on Machine Learning}, pages
  3652--3661. PMLR, 2019.

\bibitem[Levine et~al.(2020)Levine, Kumar, Tucker, and Fu]{levine2020offline}
Sergey Levine, Aviral Kumar, George Tucker, and Justin Fu.
\newblock Offline reinforcement learning: Tutorial, review, and perspectives on
  open problems.
\newblock \emph{arXiv preprint arXiv:2005.01643}, 2020.

\bibitem[Lillicrap et~al.(2015)Lillicrap, Hunt, Pritzel, Heess, Erez, Tassa,
  Silver, and Wierstra]{lillicrap2015continuous}
Timothy~P Lillicrap, Jonathan~J Hunt, Alexander Pritzel, Nicolas Heess, Tom
  Erez, Yuval Tassa, David Silver, and Daan Wierstra.
\newblock Continuous control with deep reinforcement learning.
\newblock \emph{arXiv preprint arXiv:1509.02971}, 2015.

\bibitem[Mnih et~al.(2015)Mnih, Kavukcuoglu, Silver, Rusu, Veness, Bellemare,
  Graves, Riedmiller, Fidjeland, Ostrovski, et~al.]{mnih2015human}
Volodymyr Mnih, Koray Kavukcuoglu, David Silver, Andrei~A Rusu, Joel Veness,
  Marc~G Bellemare, Alex Graves, Martin Riedmiller, Andreas~K Fidjeland, Georg
  Ostrovski, et~al.
\newblock Human-level control through deep reinforcement learning.
\newblock \emph{Nature}, 518\penalty0 (7540):\penalty0 529, 2015.

\bibitem[Munos et~al.(2016)Munos, Stepleton, Harutyunyan, and
  Bellemare]{munos2016safe}
R{\'e}mi Munos, Tom Stepleton, Anna Harutyunyan, and Marc Bellemare.
\newblock Safe and efficient off-policy reinforcement learning.
\newblock In \emph{Advances in Neural Information Processing Systems}, pages
  1054--1062, 2016.

\bibitem[Nachum et~al.(2019)Nachum, Dai, Kostrikov, Chow, Li, and
  Schuurmans]{nachum2019algaedice}
Ofir Nachum, Bo~Dai, Ilya Kostrikov, Yinlam Chow, Lihong Li, and Dale
  Schuurmans.
\newblock Algaedice: Policy gradient from arbitrary experience.
\newblock \emph{arXiv preprint arXiv:1912.02074}, 2019.

\bibitem[Nadjahi et~al.(2019)Nadjahi, Laroche, and des Combes]{nadjahi2019safe}
Kimia Nadjahi, Romain Laroche, and Rémi~Tachet des Combes.
\newblock Safe policy improvement with soft baseline bootstrapping, 2019.

\bibitem[Paine et~al.(2020)Paine, Paduraru, Michi, Gulcehre, Zolna, Novikov,
  Wang, and de~Freitas]{paine2020hyperparameter}
Tom~Le Paine, Cosmin Paduraru, Andrea Michi, Caglar Gulcehre, Konrad Zolna,
  Alexander Novikov, Ziyu Wang, and Nando de~Freitas.
\newblock Hyperparameter selection for offline reinforcement learning, 2020.

\bibitem[Peng et~al.(2019)Peng, Kumar, Zhang, and Levine]{peng2019advantage}
Xue~Bin Peng, Aviral Kumar, Grace Zhang, and Sergey Levine.
\newblock Advantage-weighted regression: Simple and scalable off-policy
  reinforcement learning.
\newblock \emph{arXiv preprint arXiv:1910.00177}, 2019.

\bibitem[Puterman and Shin(1978)]{puterman1978modified}
Martin~L Puterman and Moon~Chirl Shin.
\newblock Modified policy iteration algorithms for discounted markov decision
  problems.
\newblock \emph{Management Science}, 24\penalty0 (11):\penalty0 1127--1137,
  1978.

\bibitem[Rajeswaran et~al.(2017)Rajeswaran, Kumar, Gupta, Vezzani, Schulman,
  Todorov, and Levine]{rajeswaran2017learning}
Aravind Rajeswaran, Vikash Kumar, Abhishek Gupta, Giulia Vezzani, John
  Schulman, Emanuel Todorov, and Sergey Levine.
\newblock Learning complex dexterous manipulation with deep reinforcement
  learning and demonstrations.
\newblock \emph{arXiv preprint arXiv:1709.10087}, 2017.

\bibitem[Scherrer et~al.(2012)Scherrer, Gabillon, Ghavamzadeh, and
  Geist]{scherrer2012approximate}
Bruno Scherrer, Victor Gabillon, Mohammad Ghavamzadeh, and Matthieu Geist.
\newblock Approximate modified policy iteration.
\newblock \emph{arXiv preprint arXiv:1205.3054}, 2012.

\bibitem[Schulman et~al.(2015)Schulman, Levine, Abbeel, Jordan, and
  Moritz]{schulman2015trust}
John Schulman, Sergey Levine, Pieter Abbeel, Michael Jordan, and Philipp
  Moritz.
\newblock Trust region policy optimization.
\newblock In \emph{International conference on machine learning}, pages
  1889--1897, 2015.

\bibitem[Siegel et~al.(2020)Siegel, Springenberg, Berkenkamp, Abdolmaleki,
  Neunert, Lampe, Hafner, Heess, and Riedmiller]{Siegel2020Keep}
Noah Siegel, Jost~Tobias Springenberg, Felix Berkenkamp, Abbas Abdolmaleki,
  Michael Neunert, Thomas Lampe, Roland Hafner, Nicolas Heess, and Martin
  Riedmiller.
\newblock Keep doing what worked: Behavior modelling priors for offline
  reinforcement learning.
\newblock In \emph{International Conference on Learning Representations}, 2020.

\bibitem[Sutton and Barto(2018)]{sutton2018reinforcement}
Richard~S Sutton and Andrew~G Barto.
\newblock \emph{Reinforcement learning: An introduction}.
\newblock MIT press, 2018.

\bibitem[Van~Hasselt et~al.(2016)Van~Hasselt, Guez, and Silver]{van2016deep}
Hado Van~Hasselt, Arthur Guez, and David Silver.
\newblock Deep reinforcement learning with double q-learning.
\newblock In \emph{Thirtieth AAAI conference on artificial intelligence}, 2016.

\bibitem[Voloshin et~al.(2019)Voloshin, Le, Jiang, and
  Yue]{voloshin2019empirical}
Cameron Voloshin, Hoang~M Le, Nan Jiang, and Yisong Yue.
\newblock Empirical study of off-policy policy evaluation for reinforcement
  learning.
\newblock \emph{arXiv preprint arXiv:1911.06854}, 2019.

\bibitem[Wang et~al.(2018)Wang, Xiong, Han, Liu, Zhang,
  et~al.]{wang2018exponentially}
Qing Wang, Jiechao Xiong, Lei Han, Han Liu, Tong Zhang, et~al.
\newblock Exponentially weighted imitation learning for batched historical
  data.
\newblock In \emph{Advances in Neural Information Processing Systems}, pages
  6288--6297, 2018.

\bibitem[Wang et~al.(2020{\natexlab{a}})Wang, Foster, and
  Kakade]{wang2020statistical}
Ruosong Wang, Dean~P. Foster, and Sham~M. Kakade.
\newblock What are the statistical limits of offline rl with linear function
  approximation?, 2020{\natexlab{a}}.

\bibitem[Wang et~al.(2021)Wang, Wu, Salakhutdinov, and
  Kakade]{wang2021instabilities}
Ruosong Wang, Yifan Wu, Ruslan Salakhutdinov, and Sham~M Kakade.
\newblock Instabilities of offline rl with pre-trained neural representation.
\newblock \emph{arXiv preprint arXiv:2103.04947}, 2021.

\bibitem[Wang et~al.(2020{\natexlab{b}})Wang, Novikov, Zolna, Merel,
  Springenberg, Reed, Shahriari, Siegel, Gulcehre, Heess,
  et~al.]{wang2020critic}
Ziyu Wang, Alexander Novikov, Konrad Zolna, Josh~S Merel, Jost~Tobias
  Springenberg, Scott~E Reed, Bobak Shahriari, Noah Siegel, Caglar Gulcehre,
  Nicolas Heess, et~al.
\newblock Critic regularized regression.
\newblock \emph{Advances in Neural Information Processing Systems}, 33,
  2020{\natexlab{b}}.

\bibitem[Wu et~al.(2019)Wu, Tucker, and Nachum]{wu2019behavior}
Yifan Wu, George Tucker, and Ofir Nachum.
\newblock Behavior regularized offline reinforcement learning, 2019.

\bibitem[Zanette(2021)]{zanette2021exponential}
Andrea Zanette.
\newblock Exponential lower bounds for batch reinforcement learning: Batch rl
  can be exponentially harder than online rl, 2021.

\end{thebibliography}

\newpage

\appendix

\section{Gridworld example where multi-step outperforms one-step}\label{sec:app_grid}

\begin{figure}[h]
    \centering
    \includegraphics[width=\textwidth]{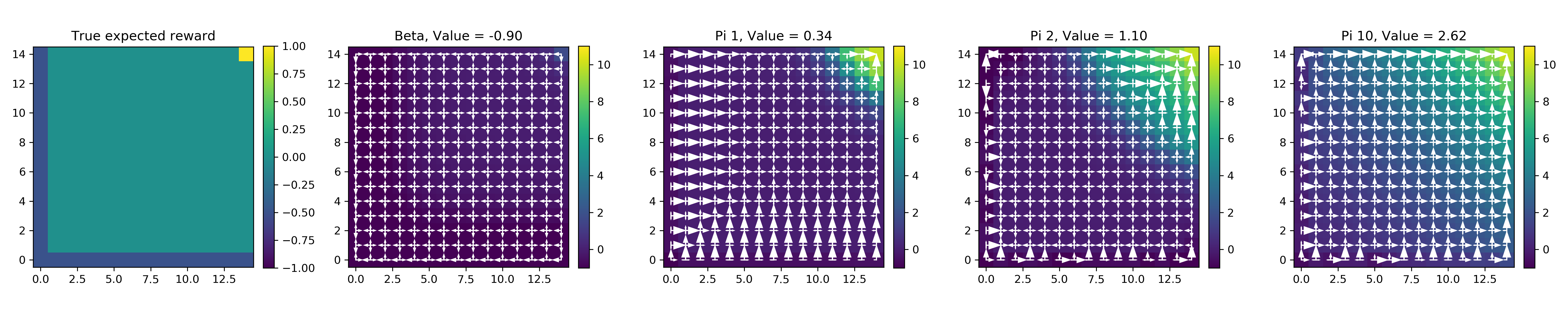}
    \includegraphics[width=\textwidth]{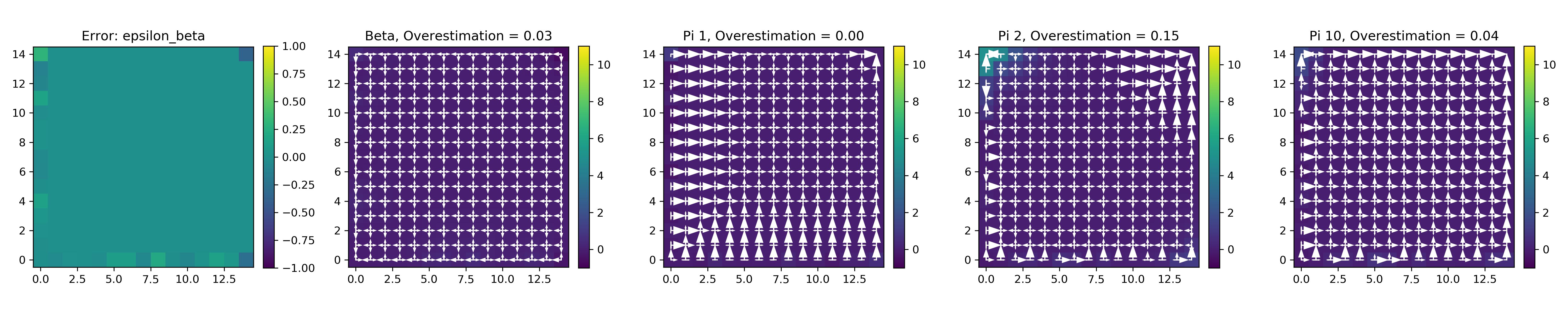}
    \caption{A gridworld example with modified behavior where multi-step is much better than one-step.}
    \label{fig:multi_gridworld}
\end{figure}

As explained in the main text, this section presents an example that is only a slight modification of the one in Figure \ref{fig:gridworld}, but where a multi-step approach is clearly preferred over just one step. The data-generating and learning processes are exactly the same (100 trajectories of length 100, discount 0.9, $ \alpha = 0.1$ for reverse KL regularization). The only difference is that rather than using a behavior that is a mixture of optimal and uniform, we use a behavior that is a mixture of maximally suboptimal and uniform. If we call the suboptimal policy $ \pi^-$ (which always goes down and left in our gridworld), then the behavior for the modified example is $ \beta = 0.2 \cdot \pi^- + 0.8 \cdot u$, where $ u $ is uniform. Results are shown in Figure \ref{fig:multi_gridworld}.

By being more likely to go to the noisy states, this behavior policy allows us to get lower variance estimates of the rewards. Essentially, the coverage of the behavior policy in this example reduces the magnitude of the evaluation errors. This allows for more aggressive planning using multi-step methods. Moreover, since the behavior is less likely to go to the good state, the behavior Q function does not propagate the signal from the rewarding state as far, harming the one-step method.

\section{Connection to policy improvement guarantees}\label{sec:app_improvement}

The regularized or constrained one-step algorithm performs an update that directly inherits guarantees from the literature on conservative policy improvement \citep{kakade2002approximately, schulman2015trust, achiam2017constrained}. These original papers consider an online setting where more data is collected at each step, but the guarantee at each step applies to our one-step offline algorithm. 

The key idea of this line of work begins with the performance difference lemma of \cite{kakade2002approximately}, and then lower bounds the amount of improvement over the behavior policy. Define the discounted state visitation distribution for a policy $ \pi$ by $ d^\pi(s) := (1-\gamma) \sum_{t=0}^\infty \gamma^t \Prob_{\rho, P, \pi}(s_t = s)$. We will also use the shorthand $ Q(s, \pi) $ to denote $ \E_{a\sim\pi|s}[Q(s,a)]$. Then we have the performance difference lemma as follows.

\begin{lemma}[Performance difference, \cite{kakade2002approximately}]
For any two policies $ \pi$ and $ \beta$,
\begin{align}
    J(\pi) - J(\beta) = \frac{1}{1-\gamma} \E_{\substack{s \sim d^\pi}}[ Q^\beta(s,\pi) - Q^\beta(s, \beta)]. %:= \frac{1}{1-\gamma} \E_{\substack{s \sim d_\pi }}[ A^\beta_\pi(s)].
\end{align}
\end{lemma}

Then, Corollary 1 from \cite{achiam2017constrained} (reproduced below) gives a guarantee for the one-step algorithm. The key idea is that when $ \pi $ is sufficiently close to $ \beta$, we can use $ Q^\beta$ as an approximation to $ Q^\pi$. 
\begin{lemma}[Conservative Policy Improvement, \cite{achiam2017constrained}]
For any two policies $ \pi$ and $ \beta$, let $ \|A^\beta_\pi\|_{\infty} = \sup_s  |Q^\beta(s,\pi) - Q^\beta(s, \beta)|$. Then,
    \begin{align}
        J(\pi) - J(\beta) \geq \frac{1}{1-\gamma} \E_{\substack{s \sim d^\beta}}\left[ \left(Q^\beta(s,\pi) - Q^\beta(s, \beta)\right) - \frac{2\gamma \|A^\beta_\pi\|_\infty}{1-\gamma} D_{TV}(\pi(\cdot|s)\|\beta(\cdot|s)) \right]
    \end{align}
where $ D_{TV}$ denotes the total variation distance.
\end{lemma}

Replacing $ Q^\beta$ with $ \widehat Q^\beta$ and the TV distance by the KL (using Pinsker's inequality), we get precisely the objective that we optimize in the one-step algorithm. This shows that the one-step algorithm indeed optimizes a lower bound on the performance difference. Of course, in practice we replace the potentially large multiplier on the divergence term by a hyperparameter, but this theory at least motivates the soundness of the approach.

We are not familiar with similar guarantees for the iterative or multi-step approaches that rely on off-policy evaluation.

\section{Experimental setup}\label{sec:app_exp_setup}

Code for our experimental setup can be found at \url{https://github.com/davidbrandfonbrener/onestep-rl}.

\subsection{Benchmark experiments (Tables \ref{tab:d4rl} and \ref{tab:multi}, Figure \ref{fig:learning_curves})}

\paragraph{Data.} We use the datasets from the D4RL benchmark \citep{fu2020d4rl}. We use the latest versions, which are v2 for the mujoco datasets and v1 for the adroit datasets.

\paragraph{Hyperparameter tuning.}
\begin{wraptable}[8]{r}{0.6\textwidth}
\vspace{-0.2cm}
    \centering
    \caption{Hyperparameter sweeps for each algorithm.}
    \begin{small}
    \begin{tabular}{lc}
        \toprule
        Algorithm & Hyperparameter set \\
        \midrule
        Reverse KL ($ \alpha$) & 
                    \{0.03, 0.1, 0.3, 1.0, 3.0, 10.0\}\\
        Easy BCQ ($ M$) & 
                    \{2, 5, 10, 20, 50, 100\}\\
        Exponentially weighted ($ \tau$) & 
                    \{0.1, 0.3, 1.0, 3.0, 10.0, 30.0\}\\
        \bottomrule
    \end{tabular}
    \end{small}
    \label{tab:hyperparams}
\end{wraptable}
We follow the practice of \cite{fu2020d4rl} and tune a small set of hyperparameters by interacting with the simulator to estimate the value of the policies learned under each hyperparameter setting. The hyperparameter sets for each algorithm can be seen in Table \ref{tab:hyperparams}. We tune hyperparameters using 3 seeds, but then evaluate the best hyperparameter by training on an additional 7 seeds and then report results on the 10 total seeds.

This may initially seem like ``cheating'', but can be a reasonable setup if we are considering applications like robotics where we can feasibly test a small number of trained policies on the real system. Also, since prior work has used this setup, it makes it easiest to compare our results if we use it too. While beyond the scope of this work, we do think that better offline model selection procedures will be crucial to make offline RL more broadly applicable. A good primer on this topic can be found in \cite{paine2020hyperparameter}.

\paragraph{Models.} All of our Q functions and policies are simple MLPs with ReLU activations and 2 hidden layers of width 1024. Our policies output a truncated normal distribution with diagonal covariance where we can get reparameterized samples by sampling from a uniform distribution and computing the differentiable inverse CDF \citep{Burkhardt2014truncated}. We found this to be more stable than the tanh of normal used by e.g. \cite{fu2020d4rl}, but to achieve similar performance when both are stable. We use these same models across all experiments.

\paragraph{One-step training procedure.} For all of our one-step algorithms, we train our $ \hat \beta $ behavior estimate by imitation learning for 500k gradient steps using Adam \citep{kingma2014adam} with learning rate 1e-4 and batch size 512. We train our $ \widehat Q^\beta$ estimator by fitted Q evaluation with a target network for 2 million gradient steps using Adam with learning rate 1e-4 and batch size 512. The target is updated softly at every step with parameter $ \tau = 0.005$. All policies are trained for 100k steps again with Adam using learning rate 1e-4 and batch size 512. 

Easy BCQ does not require training a policy network and just uses $ \hat \beta$ and $ \widehat Q^\beta$ to define it's policy. For the exponentially weighted algorithm, we clip the weights at 100 to prevent numerical instability. To estimate reverse KL at some state we use 10 samples from the current policy and the density defined by our estimated $ \hat \beta$.

Each random seed retrains all three models (behavior, Q, policy) from different initializations. We use three random seeds.

\paragraph{Multi-step training procedure.} For multi-step algorithms we use all the same hyperparameters as one-step. We initialize our policy and Q function from the same pre-trained $ \hat \beta$ and $ \widehat Q^\beta$ as we use for the one-step algorithm trained for 500k and 2 million steps respectively. Then we consider 5 policy steps. To ensure that we use the same number of gradient updates on the policy, each step consists of 20k gradient steps on the policy followed by 200k gradient steps on the Q function. Thus, we take the same 100k gradient steps on the policy network. Now the Q updates are off-policy so the next action $a'$ is sampled from the current policy $ \pi_i$ rather than from the dataset. 

\paragraph{Iterative training procedure.} For iterative algorithms we again use all the same hyperparameters and initialize from the same $ \hat \beta$ and $ \widehat Q^\beta$. We again take the same 100k gradient steps on the policy network. For each step on the policy network we take 2 off-policy gradient steps on the Q network.

\paragraph{Evaluation procedure.} To evaluate each policy we run 100 trajectories in the environment and compute the mean. We then report the mean and standard error over 10 training seeds.

\subsection{MSE experiment (Figure 3)}

\paragraph{Data.} To get an independently sampled dataset of the same size as the training set, we use the behavior cloned policy $ \hat \beta$ to sample 1000 trajectories. The checkpointed policies are taken at intervals of 5000 gradient steps from each of the three training seeds. 

\paragraph{Training procedure.} The $\widehat Q^{\pi_i}$ training procedure is the same as before so we use Adam with step size 1e-4 and batch size 512 and a target network with soft updates with parameter 0.005. We train for 1 million steps. 

\paragraph{Evaluation procedure.} To evaluate MSE, we sample 1000 state, action pairs from the original training set and from each state, action pair we run 3 rollouts. We take the mean over the rollouts and then compute squared error at each state, action pair and finally get MSE by taking the mean over state, action pairs. The reported reverse KL is evaluated by samples during training. At each state in a batch we take 10 samples to estimate the KL at that state and then take the mean over the batch.

\subsection{Gridworld experiment (Figure 4)}

\paragraph{Environment.} The environment is a 15 x 15 gridworld with deterministic transitions. The rewards are deterministically 1 for all actions taken from the state in the top right corner and stochastic with distribution $ \mathcal{N}(-0.5, 1)$ for all actions taken from states on the left or bottom walls. The initial state is uniformly random. The discount is 0.9.

\paragraph{Data.} We collect data from a behavior policy that is a mixture of the uniform policy (with probability 0.8) and an optimal policy (with probability 0.2). We collect 100 trajectories of length 100.

\paragraph{Training procedure.} We give the agent access to the deterministic transitions. The only thing for the agent to do is estimate the rewards from the data and then learn in the empirical MDP. We perform tabular Q evaluation by dynamic programming. We initialize with the empirical rewards and do 100 steps of dynamic programming with discount 0.9. Regularized policy updates are solved for exactly by setting $ \pi_i(a|s) \propto \beta(a|s) \exp(\frac{1}{\alpha} \widehat Q^{\pi_{i-1}}(s,a))$.

\subsection{Overestimation experiment (Figure 5)}

This experiment uses the same setup as the MSE experiment. The main difference is we also consider the Q functions learned during training and demonstrate the overestimation relative to the Q functions trained on the evaluation dataset as in the MSE experiment.

\subsection{Mixed data experiment (Figure 6)}

We construct datasets with $ p_m = \{0.0, 0.1, 0.2, 0.4, 0.6, 0.8, 1.0\}$ by mixing the random and medium datasets from D4RL and then run the same training procedure as we did for the benchmark experiments. Each dataset has the same size, but a different proportion of trajectories from the medium policy.

\newpage
\section{Learning curves}\label{sec:app_extra_exp}

In this section we reproduce the learning curves and hyperparameter plots across the one-step, multi-step, and iterative algorithms with reverse KL regularization, as in Figure \ref{fig:learning_curves}. 

\vspace{-0.2cm}
\begin{figure}[h]
    \centering
    \includegraphics[width=0.85\textwidth]{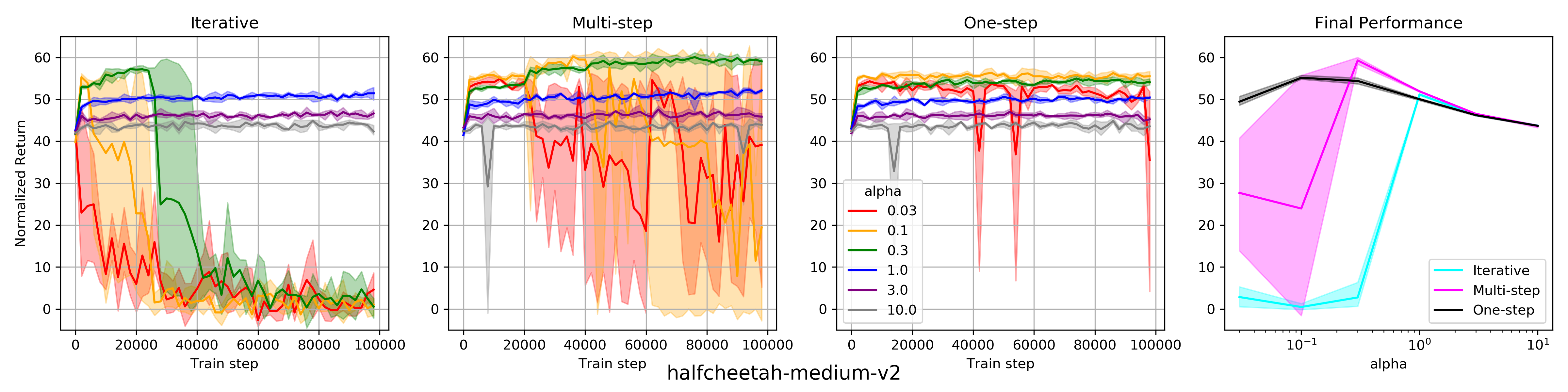}
    \includegraphics[width=0.85\textwidth]{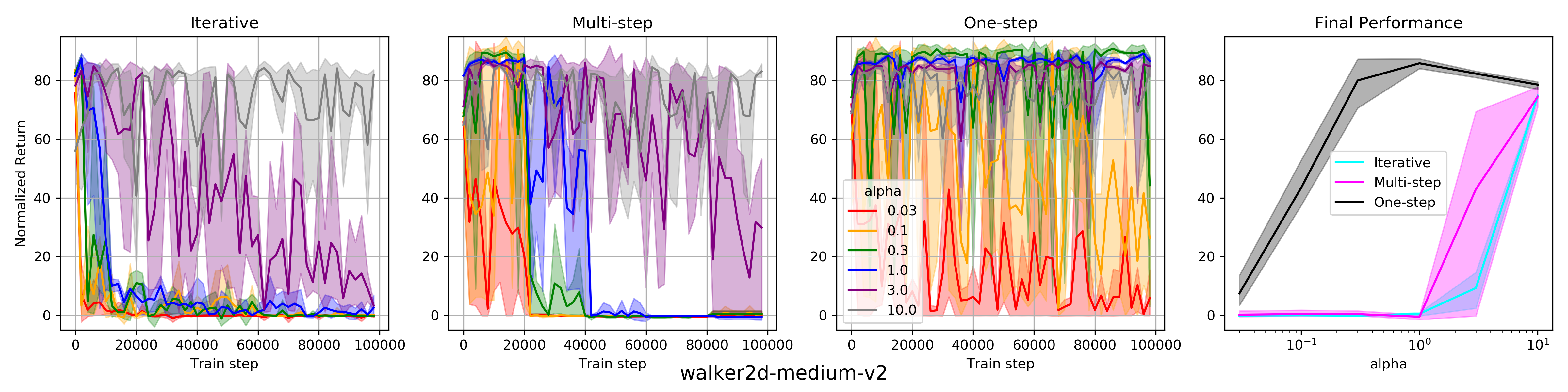}
    \includegraphics[width=0.85\textwidth]{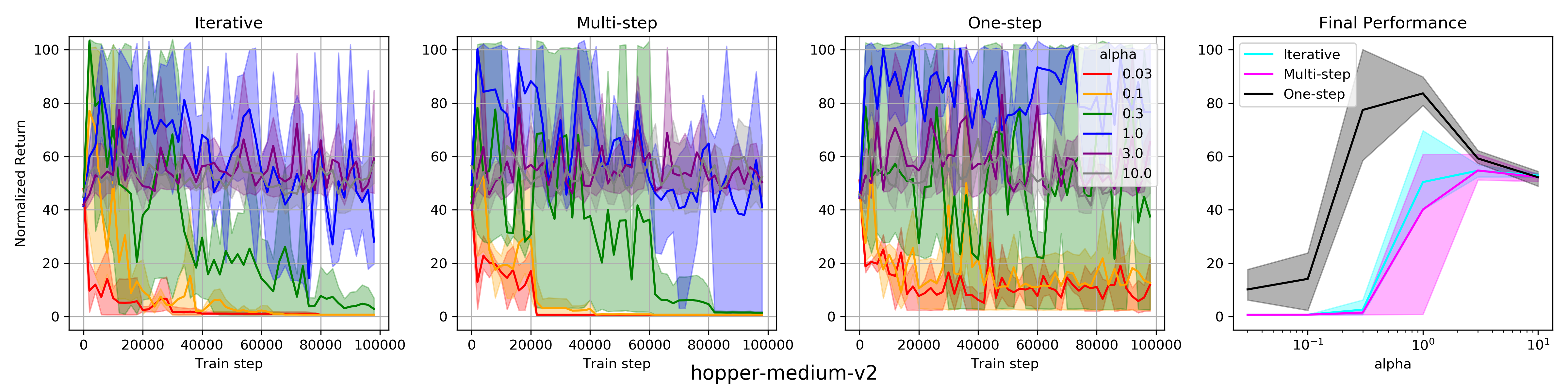}
    \vspace{-0.2cm}
    \caption{Learning curves on the medium datasets.}
    \label{fig:app_lc_medium}
\end{figure}

\begin{figure}[h]
    \centering
    \includegraphics[width=0.85\textwidth]{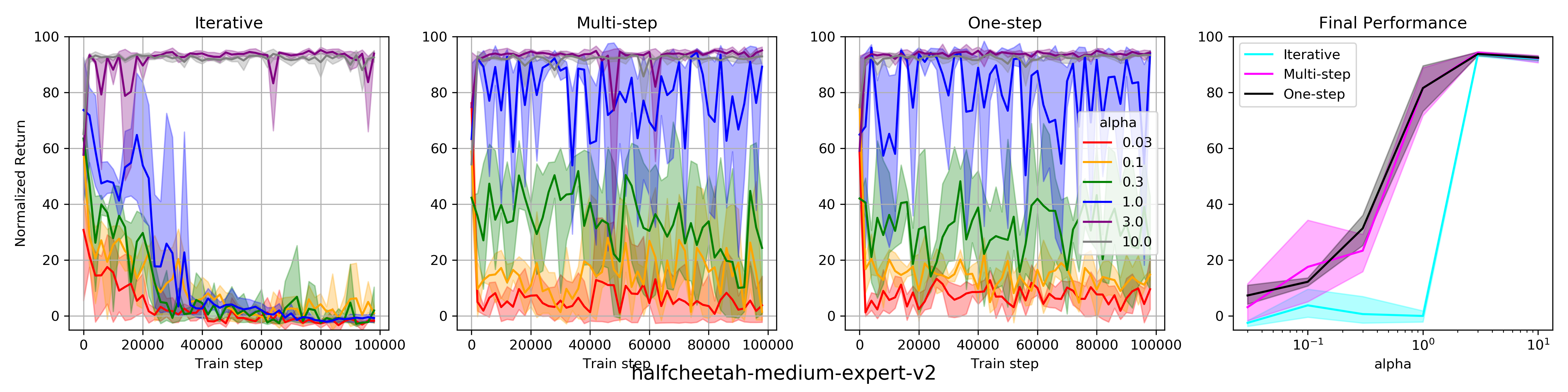}
    \includegraphics[width=0.85\textwidth]{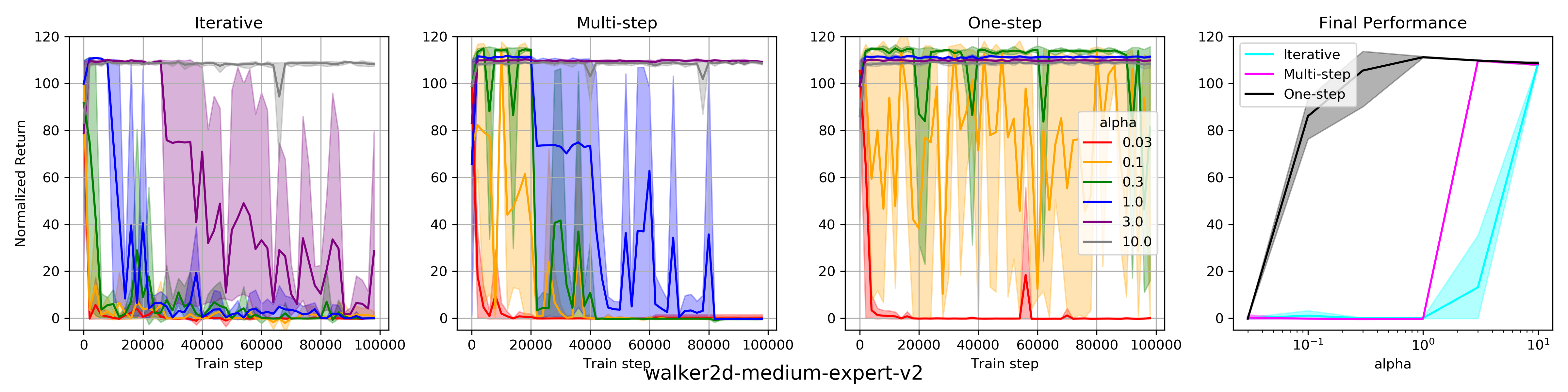}
    \includegraphics[width=0.85\textwidth]{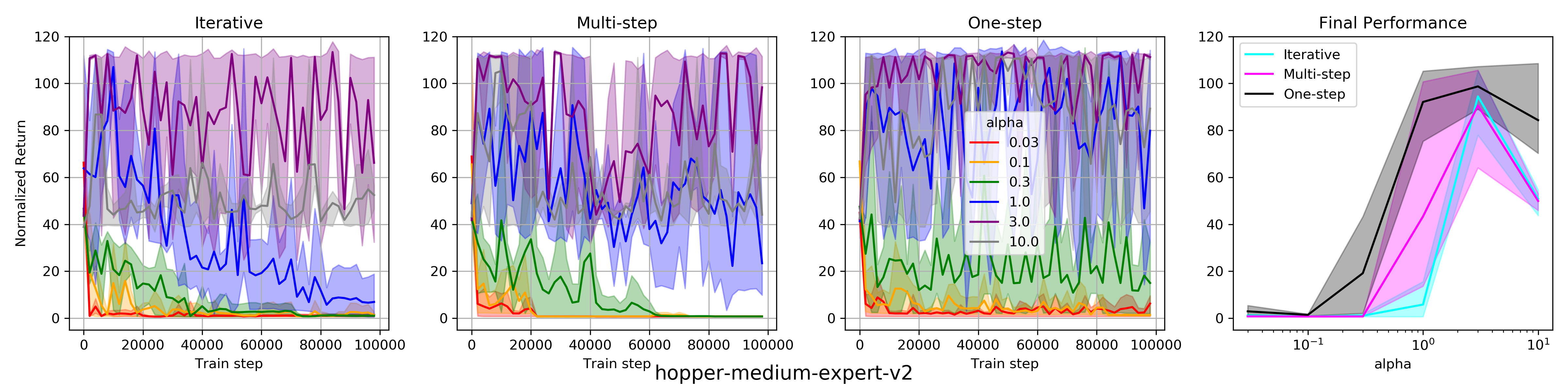}
    \vspace{-0.2cm}
    \caption{Learning curves on the medium-expert datasets.}
    \label{fig:app_lc_medium-expert}
\end{figure}

\begin{figure}[h]
    \centering
    \includegraphics[width=0.85\textwidth]{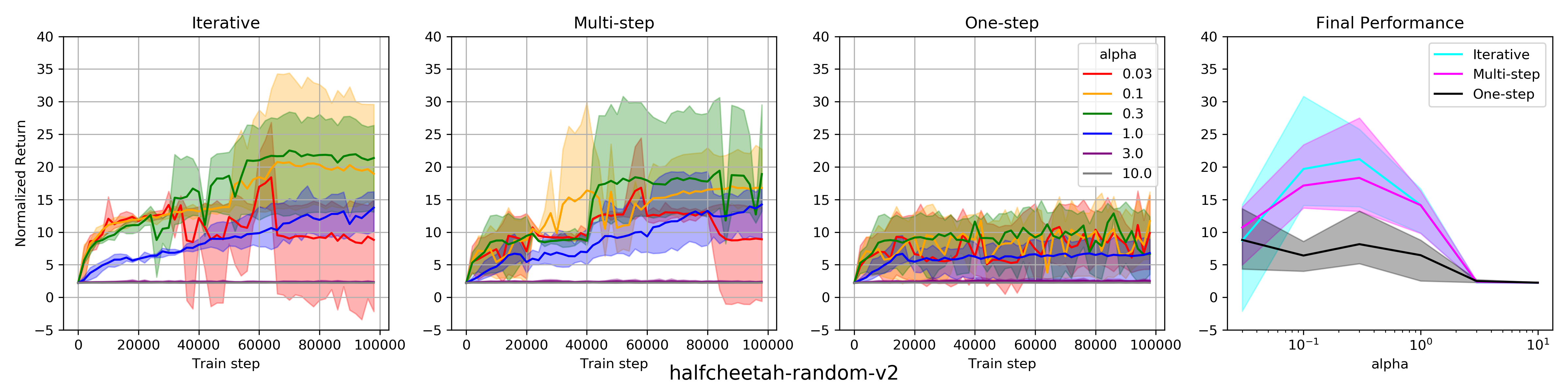}
    \includegraphics[width=0.85\textwidth]{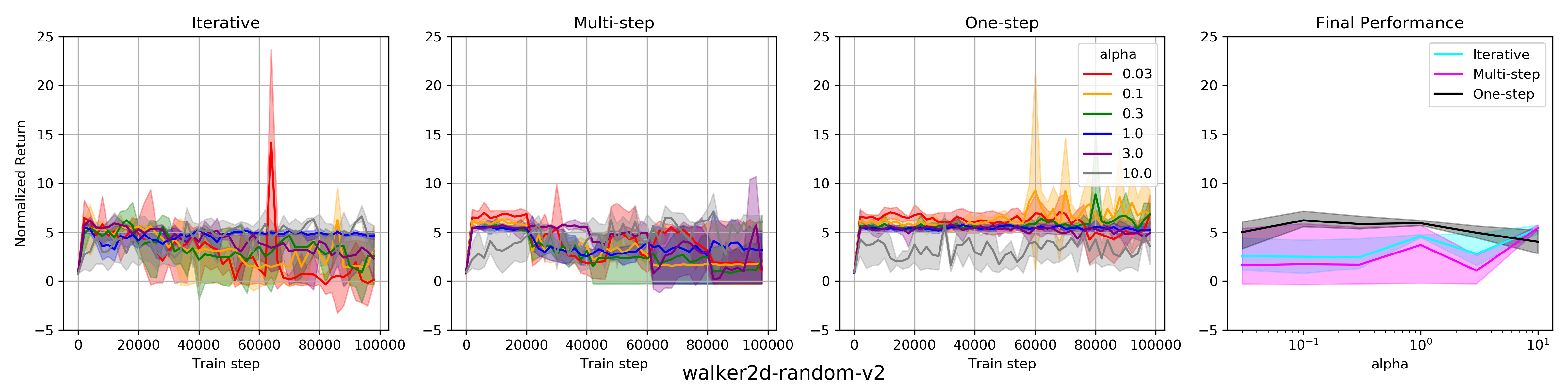}
    \includegraphics[width=0.85\textwidth]{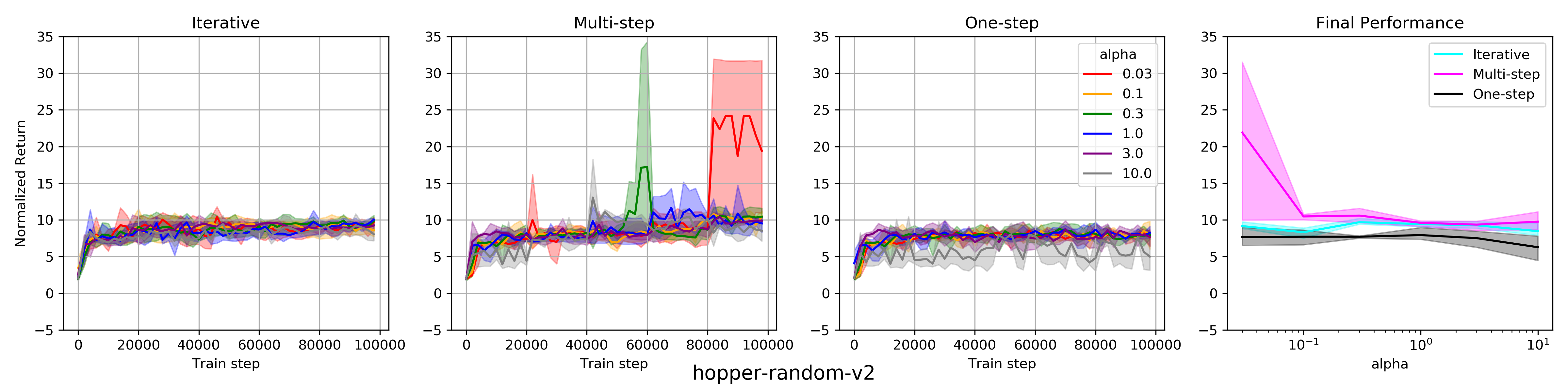}
    \vspace{-0.2cm}
    \caption{Learning curves on the random datasets.}
    \label{fig:app_lc_random}
\end{figure}

\end{document}